\documentclass[letterpaper, 10 pt, conference]{ieeeconf}  % Comment this line out if you need a4paper

\IEEEoverridecommandlockouts                              % This command is only needed if
\makeatletter
\let\NAT@parse\undefined
\makeatother
\usepackage[numbers,sort&compress]{natbib}

\usepackage[pdftex]{graphicx}
\usepackage{amsmath}
\usepackage{amssymb}  % assumes amsmath package installed
\usepackage{subcaption}
\usepackage{multirow}
\usepackage{array,booktabs}
\usepackage{diagbox}
\usepackage{balance}
\usepackage{xspace}
\usepackage{algorithm}
\usepackage{algpseudocode}
\usepackage{adjustbox}
\usepackage{romannum}
\usepackage{mathtools}

\usepackage[final]{hyperref}
\hypersetup{
 colorlinks=true,
 linkcolor=blue,
 filecolor=magenta,
 urlcolor=blue,
 citecolor=black
}
\usepackage{cleveref}
\crefname{figure}{Fig.}{Figs.}
\Crefname{figure}{Fig.}{Figs.}
\usepackage[font=small]{caption}
\usepackage{rpm_SIunits}
\usepackage{rpm_acronyms}
\usepackage{rpm_math}
\usepackage{rpm_misc}
\usepackage{bbm}

\usepackage{textcomp}

\usepackage{soul,color}
\usepackage{lipsum}
\usepackage{dsfont}

\usepackage{xcolor}
\usepackage{colortbl}

\usepackage{amssymb}% http://ctan.org/pkg/amssymb
\usepackage{pifont}% http://ctan.org/pkg/pifont
\usepackage{tikz} % \checkmark
\usepackage{subcaption}
\usepackage{caption}
\title{\LARGE \bf Informative Object-centric Next Best View \\for Object-aware 3D Gaussian Splatting in Cluttered Scenes
}     

\author{
\makebox[\textwidth][c]{%
    Seunghoon Jeong${}^{1}$, 
    Eunho Lee${}^{1}$, 
    Jeongyun Kim${}^{2}$, and 
    Ayoung Kim${}^{2*}$
}%
\thanks{$^\dagger$This work was partly supported by Hyundai Motor Company and Kia, and the Institute of Information \& communications Technology Planning \& Evaluation (IITP) grant funded by the Korea government(MSIT) No.2022-0-00480, Development of Training and Inference Methods for Goal-Oriented Artificial Intelligence Agents and the Korea government(MSIT) [NO.RS-2021-II211343, Artificial Intelligence Graduate School Program (Seoul National University)].}
\thanks{$^{1}$S. Jeong and E. Lee are with the Interdisciplinary Program in Artificial Intelligence, SNU, Seoul, S. Korea {\tt\small [shoon0602, eunho1124]@snu.ac.kr}}%
\thanks{$^{2}$J. Kim and A. Kim are with the Dept. of Mechanical Engineering, SNU, Seoul, S. Korea {\tt\small [jeongyunkim, ayoungk]@snu.ac.kr}}%
}

\begin{document}

%\onecolumn
\maketitle
\thispagestyle{empty}
\pagestyle{empty}

\begin{abstract}

In cluttered scenes with inevitable occlusions and incomplete observations, selecting informative viewpoints is essential for building a reliable representation.
% Since occlusions and incomplete observations are inevitable in cluttered scenes, a robot must select informative viewpoints to build a reliable representation of the environment. 
% In this context, \ac{3DGS} offers a distinct advantage, as it can refine the representation with new observations while explicitly guiding the selection of subsequent viewpoints.
In this context, \ac{3DGS} offers a distinct advantage, as it can explicitly guide the selection of subsequent viewpoints and then refine the representation with new observations.
% Integration of \ac{NBV} and 3D reconstruction allows efficient viewpoint selection and fast scene updates, benefiting robotic manipulation in cluttered environments.
% Integration of \ac{NBV} and 3D reconstruction allows efficient identification of informative viewpoints, benefiting robotic manipulation in cluttered environments.
% Integration of \ac{NBV} and 3D reconstruction enables efficient identification of informative viewpoints and rapid incorporation of newly acquired observations into a scene representation, which is particularly beneficial for robotic manipulation in cluttered environments.
% However, existing approaches often overlook semantic information that is crucial for manipulation and tend to prioritize exploitation over exploration, limiting their ability to complete object-level reconstructions.
However, existing approaches rely solely on geometric cues, neglect manipulation-relevant semantics, and tend to prioritize exploitation over exploration.
To tackle these limitations, we introduce an instance-aware \ac{NBV} policy that prioritizes underexplored regions by leveraging object features.
Specifically, our object-aware \ac{3DGS} distills instance-level information into one-hot object vectors, which are used to compute confidence-weighted information gain that guides the identification of regions associated with erroneous and uncertain Gaussians.
% To advance \ac{NBV} selection, we introduce a policy that explicitly prioritizes underexplored regions. This policy is enabled by a semantic 3D Gaussian Splatting framework, which distills instance-level information into one-hot object vectors and employs confidence-aware information gain to guide the NBV process with greater accuracy and robustness.
% To address these limitations, we propose a semantic 3D Gaussian Splatting framework that distills instance-level information into one-hot object vector and leverages it to guide an NBV policy toward underexplored regions. 
% Furthermore, our method can be easily adapted to an object-centric \ac{NBV}, which focuses view selection on a target object and its surrounding context.
% This enables the robot to concentrate its scanning on task-relevant areas instead of the entire scene, thereby improving reconstruction robustness to object placement.
Furthermore, our method can be easily adapted to an object-centric \ac{NBV}, which focuses view selection on a target object, thereby improving reconstruction robustness to object placement.
% This demonstrates significant effectiveness in allowing the robot to focus scanning solely on task-relevant areas, thus accelerating execution.
Experiments demonstrate that our \ac{NBV} policy reduces depth error by up to 77.14\% on the synthetic dataset and 34.10\% on the real-world GraspNet dataset compared to baselines. Moreover, compared to targeting the entire scene, performing \ac{NBV} on a specific object yields an additional reduction of 25.60\% in depth error for that object. We further validate the effectiveness of our approach through real-world robotic manipulation tasks.
\end{abstract}

% 새로운 정보를 많이 얻을 수 있는 뷰를 추정하고 얻은 뷰를 빠르게 지식에 반영할 수 있는 next best view 기술은 특히 cluttered scene에서의 manipulation에 효과적이다. 그러나 현존하는 연구는 manipulation에 유용한 semantic 정보를 활용하지 않을 뿐 아니라 물체 reconstruction을 완성하기 위한 exploration보다 exploitation에 집중한다. 

% 이를 위해 우리는 instance 정보를 one-hot object vector로 distill한 semantic 3dgs와 그를 이용하여 못 봤던 부분에 더 집중하는 next best view method를 제안한다. 또한, 간단한 변경을 통해 특정 물체와 그 주변에 집중한 next best view도 가능하다. 이는 로봇이 필요한 부분만을 스캔하여 빠르게 작업을 수행하는 데 매우 효과적이다.

% 우리는 여러 synthetic과 real dataset에서 이 nbv가 depth 오차를 00% 감소시켜줌을 확인하였고, 실제 robot experiment에서도 활용하여 manipulation task를 수행하였다.
\section{Introduction}
\label{sec:intro}

\definecolor{mygreen}{HTML}{008000}
\begin{figure}[!t]
  \centering
  \includegraphics[width=1\columnwidth, trim=0cm 0cm 0cm 0cm, clip]{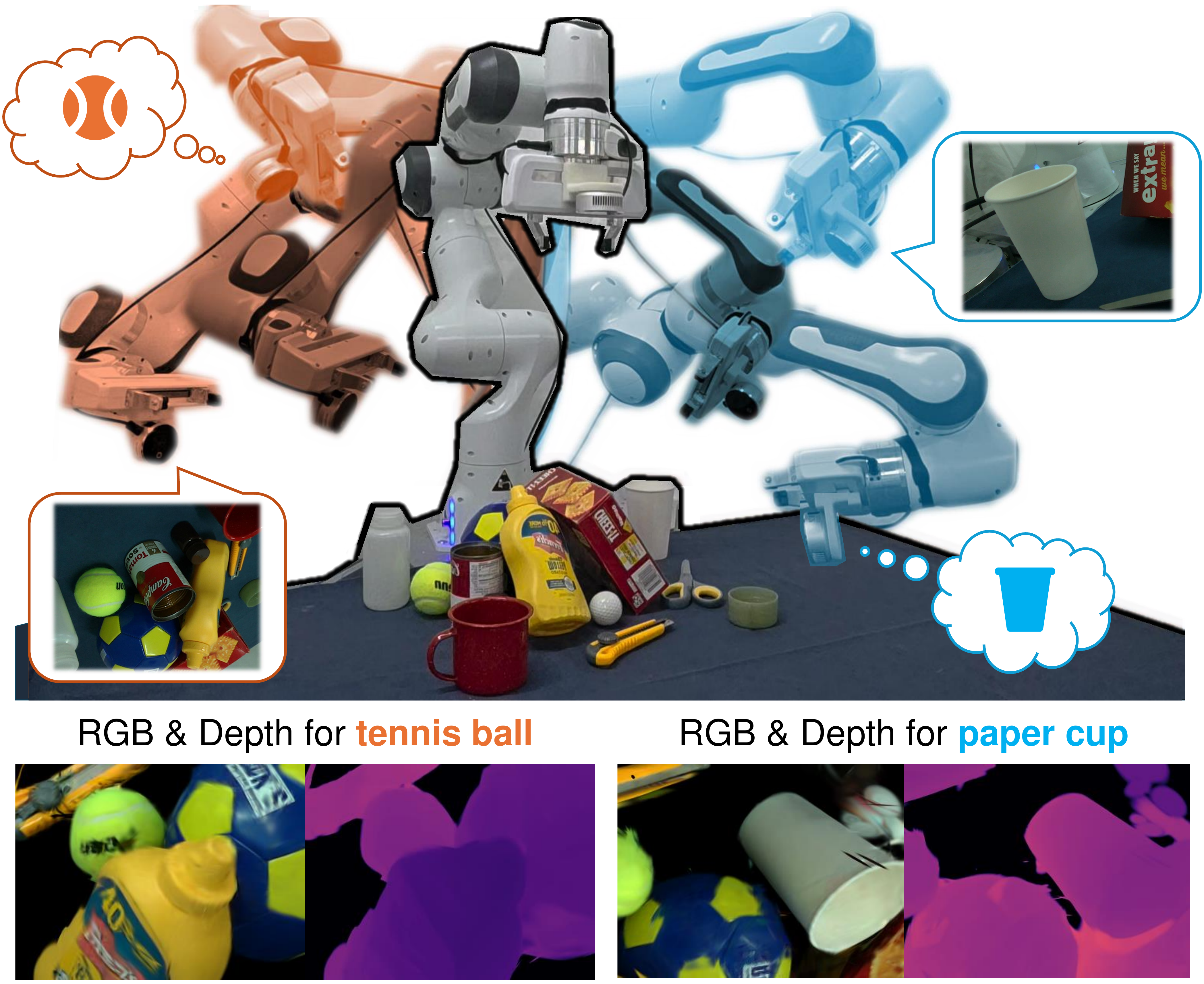}
  % \fbox{\rule{0pt}{1.0\columnwidth} \rule{0.9\columnwidth}{0pt}}
  \caption{Object-centric \ac{NBV} for manipulation in cluttered scenes. Instead of reconstructing the entire scene, we prioritize task-relevant regions surrounding the target object (tennis ball (\textcolor{red}{red}) or cup (\textcolor{blue}{blue})). Using these selected views with our object-aware \ac{3DGS} yields improved novel view synthesis and depth reconstruction, as shown below.}
  % cluttered scene에서 manipulation을 할 때 전체 scene을 reconsturction 하는 것보다 target object 위주 (tennis ball for green, cup for blue) 를 우선적으로 reconstruct하는 것이 효율적이다. 우리는 이를 위해 물체를 구분할 수 있는 semantic 3DGS와 거기서 얻을 수 있는 information을 기반으로 Next best view를 결정하는 policy를 사용한다.
  \label{fig:main}
  \vspace{-6mm}
\end{figure}

Accurate perception of the operating environment including the geometry of objects is essential for successful robotic manipulation. While many prior works \cite{fang2020graspnet, wu2024economic} estimate grasping points from a single view depth image, real-world cluttered scenes often contain stacked or irregularly distributed objects, where severe occlusions occur. In such cases, predefined viewpoints may either lack necessary information or include redundant observations, making them insufficient and ineffective for reliable manipulation \cite{murali20206, newbury2023deep}.  This motivates the need for a robot to \textit{actively} move and select informative viewpoints based on the scene context or the target object and incrementally refine its understanding of the scene from the newly captured perception.

% 로봇이 manipulation task를 제대로 수행하기 위해선 잡고자 하는 물체의 geometry를 포함한 surrounding environment를 정확하게 인지해야 한다. 기존에 보통 Bird's eye view에서의 single view depth image를 통해 물체의 grasping point를 얻는 연구들이 많이 진행되어 왔지만, 실제 practical한 usage에서 로봇은 여러 물체가 쌓여있거나 불규칙적으로 분포되어 있는 cluttered scene을 마주하게 된다. 이런 상황에서는 물체들간의 occlusion이 크게 작용하기 때문에 소수의 pre-define된 위치의 view만으로는 manipulation을 제대로 수행할 수 없다. 따라서 로봇에게는 scene의 상황이나 목표로 하는 물체에 따라 더 많은 정보를 줄 수 있는 view를 선택하고 그를 이용하여 점진적으로 scene을 이해해나갈 수 있는 능력이 필요하다.

Identifying a minimal set of training views that preserves reconstruction quality frames the problem of view selection and \acf{NBV} planning. Voxel-based methods \cite{zhang2023affordance, jin2025activegs} estimate volumetric completeness via per-voxel occupancy, but require high memory at finer resolution and do not account for photometric quality. Bayes’ Rays \cite{goli2024bayes} addresses this limitation by parameterizing perturbations to \ac{NeRF} \cite{mildenhall2021nerf} parameters to quantify uncertainty, yet its slow rendering speed makes it unsuitable for robotic tasks. 

% Identifying a minimal set of training views that preserves reconstruction quality frames the problem of view selection and \ac{NBV} planning. Using such an algorithm, the information gain of multiple candidate views can be quantified, and the view with the highest gain can be added to the training set to progressively complete the scene. Methods based on voxel representations \cite{zhang2023affordance, jin2025activegs, nagami2025vista} typically estimate volumetric completeness by measuring per-voxel occupancy, but this approach suffers from high memory usage depending on voxel resolution and does not account for photometric quality. Bayes’ Rays \cite{goli2024bayes} addresses this limitation by parameterizing perturbations to \ac{NeRF} \cite{mildenhall2021nerf} parameters to quantify uncertainty, yet its slow rendering speed makes it unsuitable for robotic tasks. 

% view selection과 Next best view planning은 reconstruction quality를 해치지 않는 최소한의 training view set을 찾는 문제로, 이 알고리즘을 이용해 여러 candidate view 들의 information gain을 quantify하고 그것이 가장 큰것을 새로운 training view로 추가하여 scene을 완성해나갈 수 있다. Voxel-based representation을 사용한 연구들은 voxel별로 occupancy 등을 이용하여 volumetric completeness를 비교하여 view를 선택하지만, voxel의 해상도에 따라 메모리 사용량이 크고 photometric quality를 고려하지 않는다. NeRF parameter의 perturbation을 파라미터화하여 uncertainty를 quantify한 Bayes' Rays는 이를 해소하지만 NeRF의 느린 rendering 속도로 인해 로봇 task에 적합하지 않다.

Compared to these approaches, methods \cite{jiang2024fisherrf, wilson2025pop, strong2024next} based on \acf{3DGS} \cite{kerbl20233d} offer the advantages of using an explicit representation, which enables per-Gaussian uncertainty estimation and faster rendering. By leveraging the Jacobian of Gaussian parameters with respect to the rendering output, these methods avoid heuristic modelings and select views in a direct and efficient manner. However, since well-observed Gaussians with low uncertainty still dominate the rendering contribution, these \ac{NBV} strategies tend to focus on exploitation rather than exploration. Moreover, these methods are designed for reconstructing the entire scene, making them less robust for manipulation tasks targeting objects that are occluded, small, or far from the center. In such cluttered environments, robots often require object-centric \ac{NBV} planning that can focus on a specific object.

% 문장 김.

% 이에 비해 3DGS-based method들은 explicit한 representation를 가져 per-Gaussian uncertainty를 구할 수 있고 rendering이 매우 빠르다는 장점을 가지고 있다. gaussian의 parameter들의 rendering에 대한 jacobian을 uncertainty qunatification에 사용하는 이 방법들은 heurisitc하지도 않고 view를 직접적이고 효율적으로 선택하지만, observation이 많아 certain한 Gaussian들이 여전히 rendering에 대한 기여도를 dominant하기 때문에 exploration 보다 exploitation에 집중하는 경향이 있다. 또한, 이 방법들은 모두 전체 scene에 대한 reconstruction을 목표로 할 뿐, 특정 물체에 집중된 NBV가 불가능하여 cluttered scene에서 가려지거나 작은 물체들에 대해서 manipulation을 수행할 때 robust 하지 않다.

To tackle these limitations, we propose an informative, object-centric \ac{NBV} policy integrated with object-aware \ac{3DGS}.
Our key insight is to inject instance-aware confidence score into uncertainty quantification so that information gain and view selection prioritize uncertain, task-relevant regions. This enables \ac{NBV} for both whole scene coverage and target-conditioned inspection with better novel view synthesis and depth reconstruction as shown in \figref{fig:main}.
Our object-aware \ac{3DGS} distills object instance masks from image foundation models \cite{liu2024grounding, kirillov2023segment, ren2024grounded} into a one-hot object vector, reconstructs scenes in fewer iterations than optimization using only RGB images and yields explicit per-Gaussian object features that guide \ac{NBV}.
We validate our approach through substantial improvements in both novel view synthesis and depth estimation, as well as successful deployment in real-world robotic arm experiments.
Our main contributions are as follows:

\begin{itemize}
    \item \textbf{Instance-aware Information Gain for Next Best View: }
    By weighting each Gaussian’s information gain with a confidence score from object features, our \ac{NBV} planner prioritizes poorly observed yet task-relevant regions, encouraging exploration.
    % object one-hot vector를 confidence로 활용하여 training view에서 rendering quality에 대한 기여도가 작은 gaussian들을 더 집중할 수 있는 information gain 식을 제시한다.
    \item \textbf{Object-aware 3D Gaussian Splatting and Fast Scene Refinement: } 
    We introduce an object-aware \ac{3DGS} that reconstructs and segments cluttered scenes in just a few iterations with low training time, by fusing incoming RGB images and object instance masks through a one-hot object vector.
    % 지속적으로 view가 추가되는 tabletop cluttered scene setting에서 rgb와 object mask를 이용해 빠르게 3d reconstruction 및 segmentation을 수행하는 sparse-view 3dgs framework을 소개한다.
    \item \textbf{Object-centric Reconstruction and Manipulation: }
    Beyond whole-scene modeling, our system can focus \ac{NBV} around user-specified objects and produces optimal 6-DoF grasp poses in cluttered scenes.
    % 전체 3d scene뿐만 아니라 robotic manipulation 시 practical한 활용을 위해 목표로 하는 물체 주변에 집중하고 최적의 6dof grasp를 결정하도록 한다.
\end{itemize}

\section{related work}
\label{sec:relatedwork}
\subsection{View Selection and Next Best View}

View selection and \ac{NBV} have received significant attention as approaches for identifying the most informative next viewpoint and selecting an optimal set of input images that effectively summarize the scene content. Previous approaches in radiance field-based \ac{NBV} have included methods that treat uncertainty directly as a learnable parameter \cite{feng2024naruto} or exploit heuristic information gain functions \cite{chen2025activegamer}. However, these methods fail to fully capture epistemic uncertainty arising from insufficient knowledge. 
% \cite{klasson2024sources}

Some approaches \cite{goli2024bayes, jiang2024fisherrf, wilson2025pop, strong2024next}, on the other hand, have proposed quantifying uncertainty using the Hessian matrix of the negative log-likelihood and successfully applied to both \ac{NeRF} and \ac{3DGS}. Specifically, FisherRF \cite{jiang2024fisherrf} directly measured uncertainty by leveraging the Jacobian of rendering outputs with respect to the Gaussian primitives in \ac{3DGS}. Subsequently, POp-GS \cite{wilson2025pop} introduced concepts from optimal experimental design \cite{kiefer1974general} into information gain calculations, partially accounting for parameter correlations. Next Best Sense \cite{strong2024next} further expanded its applicability by incorporating depth information into the information gain.

% 기존 연구에서는 radiance field에서 next best view를 수행하기 위해 uncertainty를 직접 learnable parameter로 설정하는 접근법도 제안되었지만, 이 방식은 지식의 부족으로 인한 epistemic uncertainty를 제대로 반영하지 못하는 한계가 있다. 반면, 최근에는 negative log likelihood의 2차 미분인 Hessian matrix를 이용하여 uncertainty를 정량적으로 분석하는 연구가 제안되었고, 이는 NeRF와 3DGS 모두에 성공적으로 적용되었다. 특히, FisherRF는 3DGS의 Gaussian primitive에 대한 rendering 결과의 Jacobian을 이용해 직접 uncertainty를 측정했다. 이후 Pop-GS는 optimal experimental design 개념을 활용하여 information gain 계산에 parameter 간의 correlation을 부분적으로 반영했으며, Next Best Sense는 depth 정보까지 information gain 계산에 포함하여 활용 가능성을 더욱 확장했다.

However, conventional information-gain-based \ac{NBV} methods often display a strong bias toward regions that have already been sufficiently observed, while leaving underexplored areas inadequately covered. This imbalance between exploration and exploitation limits their effectiveness in completing the scene understanding. We address this by incorporating per-Gaussian confidence weighting to guide viewpoint selection more evenly.

% However, conventional information-gain-based \ac{NBV} methods tend to overly focus on regions already sufficiently observed, neglecting areas that genuinely require further exploration. To mitigate this issue, our method employs per-Gaussian confidence weights derived from each Gaussian's one-hot object vector, ensuring that regions with limited observations receive higher priority. This approach effectively balances exploration and exploitation during viewpoint selection.

% 그러나 기존의 information gain 기반 NBV 방법은 이미 충분히 관찰되어 더 이상 추가적인 관찰이 필요하지 않은 영역에 계속 집중하는 경향이 있다. 이는 잘 관찰된 Gaussians들이 렌더링 품질에 크게 기여하기 때문에 information gain이 높게 평가되어, 정작 occluded region에 대한 exploration이 저해되는 문제를 야기한다. 이러한 문제를 해결하기 위해 본 연구에서는 Gaussian별로 관찰의 confidence를 명시적으로 고려하여 confidence-aware information gain을 제안한다. 이는 기존 NBV 방식과 달리 Gaussian의 object assignment 정보에서 유도된 confidence weight를 이용하여, 관찰이 부족한 영역의 Gaussians에 더 높은 가중치를 부여하고 exploration과 exploitation의 균형을 효과적으로 유지할 수 있도록 한다.

\subsection{Feature Embedded 3D Representation for Manipulation}

% Owing to its explicit representation and ease of customization, \ac{3DGS} has emerged as a powerful framework for comprehensive 3D scene understanding, spanning from low-level geometric reconstruction to high-level object localization and identification.

Recent studies \cite{kobayashi2022decomposing, kerr2023lerf} have demonstrated that applying feature distillation to volumetric rendering can jointly capture semantic and geometric information, making it highly suitable for robotic manipulation. This idea has also been extended to \ac{3DGS}, where several works \cite{qiu2024feature, qin2024langsplat} have customized the CUDA rasterizer to incorporate features from various language-based models such as CLIP \cite{radford2021learning}, SAM \cite{kirillov2023segment}, and GroundingDINO \cite{liu2024grounding}. GraspSplats \cite{ji2024graspsplats} leverages masks obtained from MobileSAM \cite{zhang2023faster} and their CLIP embeddings to design a hierarchical feature representation, enabling zero-shot object manipulation. However, directly injecting such features into Gaussians is challenging because their high dimensionality complicates optimization, slows processing, and makes it difficult to distinguish among semantically similar instances.
% However, directly injecting such features into Gaussians poses significant challenges, as the high dimensionality of the features complicates optimization, slows down overall processing, and makes it difficult to distinguish between instances sharing the same semantics.

% DFF와 LERF의 insight에서 출발하여 최근 연구자들은 feature distillation을 적용시킨 volumetric rendering이 semantic 정보와 geometry 정보를 함께 주기 때문에 robotic manipulation에 매우 적합하다는 것을 알아내었다. 이는 3DGS에도 적용되어 여러 연구에서 cuda rasterizer를 custom하여 clip, sam, groundingdino 등 여러 language-based model들의 feature를 넣어 이를 구현하였다. GraspSplats에서는 mobilesam으로 얻은 mask와 그들의 clip embedding으로 hierarchical feature를 설계하여 zero-shot object manipulation을 선보였다. 그러나, gaussian 들에 직접 이 feature를 inject하는 것은 feature의 dimension이 커서 optimzie가 어려울 뿐만 아니라 전반적인 속도를 늦추고, 같은 semantic의 instance들을 구분하기 어렵게 한다.

Accordingly, research has also emerged focusing on faster reconstruction using simpler features. Gaussian Grouping \cite{ye2024gaussian} introduced identity encodings to cluster Gaussians belonging to the same object, while ObjectGS \cite{zhu2025objectgs} and TRAN-D \cite{kim2025tran} segmented and tracked objects using Grounded SAM \cite{ren2024grounded} and then optimized a one-hot vector for each object. Similarly, we employ a one-hot vector as supervision for objects, and we further extend its use by leveraging it as a confidence measure integrated into NBV planning. This also serves as the cornerstone of our object-centric reconstruction.

% 이에 따라 좀 더 간단한 feature로 빠르게 reconstruction을 하기 위한 연구들도 나왔다. Gaussian Grouping에서는 introduced identity encodings to cluster Gaussians belonging to the same object, and objectgs와 TRAN-D는 grounded sam으로 물체들을 segment 및 tracking 한 뒤 object별의 one hot vector를 optimize했다. 우리 또한 object의 one hot vector를 supervision으로 사용하지만, 우리는 거기에서 더 나아가 이를 confidence로 활용하여 NBV planning과 결합한다.

\section{Method}

% \begin{figure*}[!t]
%   \centering
%   \fbox{\rule{0pt}{0.3\textwidth} \rule{0.9\textwidth}{0pt}}
%   \caption{Figure 2}
%   \label{fig:fig2}
%   \vspace{-4mm}
% \end{figure*}

In this section, we describe our proposed \ac{NBV} pipeline for cluttered scenes, structured into three main components. First, we introduce object-aware 3D Gaussian Splatting, which gradually reconstructs the scene as new views are incorporated. Second, we explain our improved uncertainty quantification and information gain strategy, tailored to effectively explore previously unseen regions of the scene. Lastly, we present our method for selecting views focused on specific target objects, along with subsequent object-centric robotic manipulation tasks.

% 이 섹션에서는 cluttered scene을 위한 제안된 manipulation pipeline을 세 가지 핵심 부분으로 나누어 설명한다. 첫째, 추가적인 view가 들어올 때마다 점진적으로 scene reconstruction을 수행하는 Active Semantic 3D Gaussian Splatting 기법을 소개한다. 둘째, 관찰되지 않은 영역을 효과적으로 탐색(exploration)하기 위해 uncertainty quantification 및 Information Gain 계산 방식을 개선한 방법을 설명한다. 마지막으로, 특정 물체에 초점을 맞추어 Next Best View를 선정하고 이를 기반으로 manipulation을 수행하는 과정을 제시한다.

\subsection{Object-aware 3D Gaussian Splatting}

\begin{figure}[!t]
  \centering
  \includegraphics[width=1\linewidth]{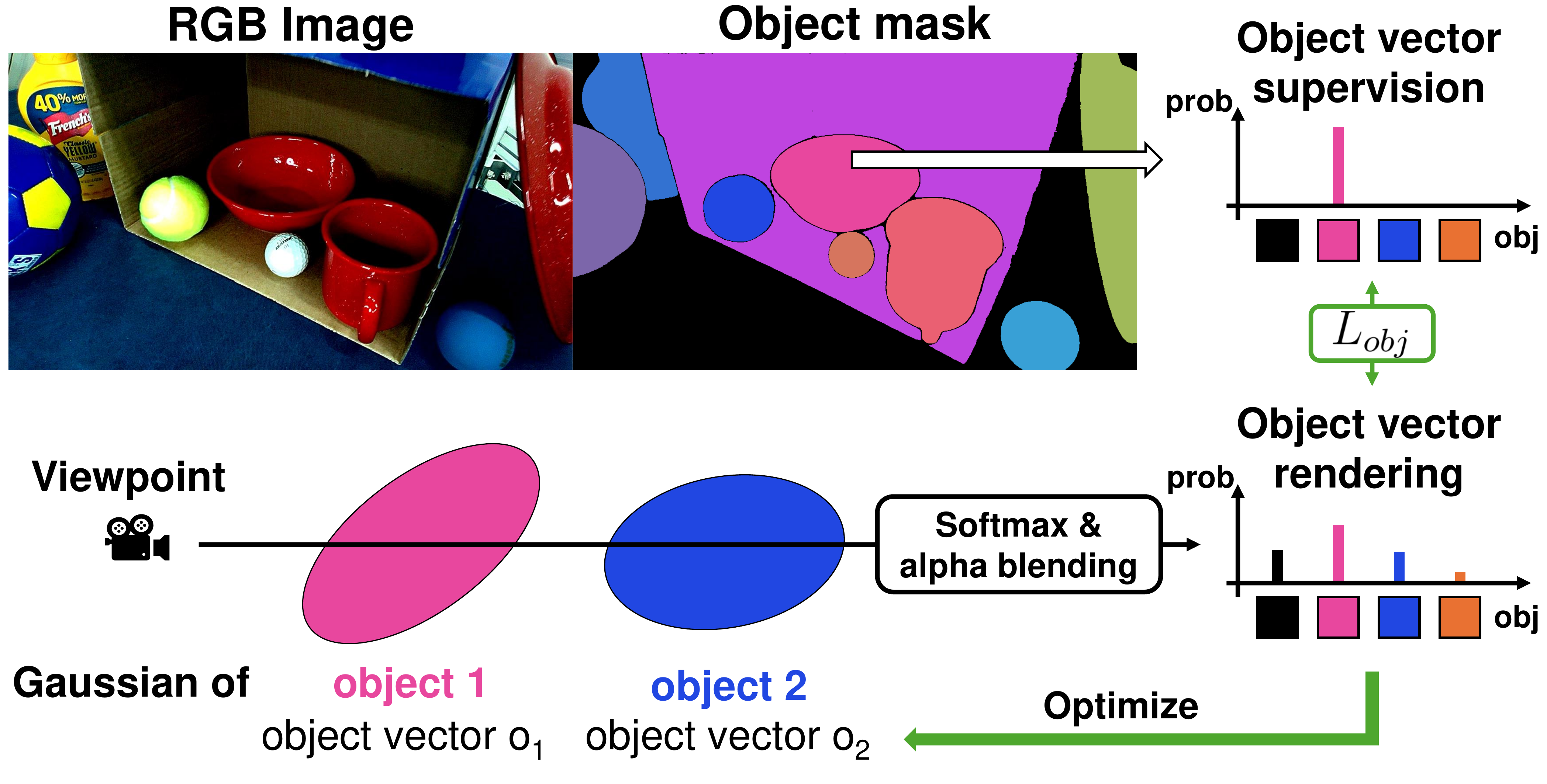}
  % \fbox{\rule{0pt}{0.5\columnwidth} \rule{0.9\columnwidth}{0pt}}
  \caption{Visualization of the rendering and learning process of the one-hot object vector. Each Gaussian stores logits for $n+1$ classes, including $n$ object categories and the background. The logits are passed through a softmax function and alpha-blended in the same manner as RGB in \ac{3DGS}. After normalization, this yields the per-pixel class probability along each ray. By supervising these probabilities with one-hot vectors obtained from instance masks, the logits for each Gaussian can be optimized.}
  % One-hot object vector의 작동과 학습 방식에 대한 visualization. gaussian마다 물체 n개와 background를 포함한 n+1개의 class에 대한 logit이 저장되어 있다. softmax된 logit을 rgb와 같은 방식으로 alpha blending을 한 뒤 합을 1로 맞춰주면 해당 ray에 해당하는 픽셀의 class probability를 얻을 수 있다. 이를 instance mask로부터 얻을 수 있는 one-hot vector와 supervision을 통해 각 gaussian의 logit을 학습할 수 있다.
  \label{fig:object vector}
  \vspace{-5mm}
\end{figure}

\subsubsection{One-hot Object Vector for Instance Segmentation}
With recent advances in 2D segmentation, obtaining object masks has become practical using image foundation models. Here, we describe how to integrate a given object mask into the \ac{3DGS} framework by adding a lightweight feature to each Gaussian. Assume the scene contains $n$ distinct objects, with object IDs assigned from $1$ to $n$ and the background labeled as $0$. Each pixel in the mask is represented as a one-hot vector of dimension $n+1$, where each channel corresponds to a specific object class or the background.
% DEVA \cite{cheng2023tracking}, 

% 2d segmentation 기술의 발전으로 DEVA, CLIP, Grounding DINO, SAM 등의 foundation model을 이용해 object mask를 쉽게 구할 수 있게 되었다. 우리는 given object mask S를 3dgs에 어떻게 inject 할 수 있는지 논하고자 한다. assume that scene에 n개의 object가 있다고 하고 우리는 그의 object id를 1~n으로 정하자. 물체가 아닌 background를 0으로 잡으면 gt mask의 각 픽셀은 n+1 dimension의 one-hot encoding vector로 나타낼 수 있다.

{\small
\begin{equation}
O' = \sum_{i} softmax(o_i) \alpha_i T_i, \quad T_{i} = \prod_{j=1}^{i-1}(1 - \alpha_j).
\end{equation}
}

\ac{3DGS} represents a 3D scene as a set of Gaussian ellipsoids, each parameterized by a center position $\mu \in \mathbb{R}^{3}$, scale $s \in \mathbb{R}^{3}$, rotation $r \in \mathbb{R}^{4}$, opacity $\sigma \in \mathbb{R}$, and spherical harmonic coefficients $c \in \mathbb{R}^{3(d+1)^2}$ where d is the degree of spherical harmonic. We assign an additional object feature vector $o_i \in \mathbb{R}^{n+1}$ to each Gaussian and optimize it by comparing the rendered results against the ground-truth object masks as shown in \figref{fig:object vector}. We apply a softmax to each $o_i$, then perform alpha-weighted blending across Gaussians to obtain a per-pixel object vector. 

{\small
\begin{equation}
O=O' + \left[1, 0, \dots, 0 \right] \cdot \left(1 - \mathbf{1}^\top O' \right)
\end{equation}
}

Since the blended object vector often sums to less than 1 due to incomplete coverage, we add the residual value to the background channel to ensure that it sums to 1. This lets each value $O_{k}$ be interpreted as the probability of the pixel belonging to object class $k$. Treating this as a multi-class segmentation, we supervise the object features with the L1 loss and the Dice loss \cite{sudre2017generalised}:

% 우리는 gaussian에 추가적인 object feature o_i를 assign하고 이를 이용하여 gt mask와 비교하며 optimize 하였다. one-hot encoding과 같이 만들기 위해 우리는 feature를 softmax 한 뒤 blending 하여 각 픽셀의 object vector를 구한다. 이 때, blending한 object vector에서 합이 1보다 작기 때문에 그만큼을 background에 더해줘서 합이 1이 되도록 하였고, 그러면 각 value들은 해당 픽셀이 특정 물체일 확률로 해석될 수 있다. 이를 multi-class segmentation으로 해석하여 우리는 l1 loss와 함께 dice loss를 사용했다.
{\small
\begin{equation}
L_{obj} = (1-\lambda_{Dice}) \times L_{1, obj} + \lambda_{Dice} \times L_{Dice},
\end{equation}
}
and the softmax of the supervised object feature yields the per-Gaussian object probability $p_i = softmax(o_i)$.

Segmentation masks primarily remove background in \ac{3DGS}. Assigning a default color to background pixels suppresses the opacity of background Gaussians, markedly reducing both the total number of Gaussians and floaters. 
% In addition, Gaussians outside the camera frustum of all training views or otherwise weakly constrained receive negligible gradients. 
In addition, we prune any Gaussian whose maximum object class probability $\max_{k}(p_{i,k})$ falls below a threshold $\delta_{obj}$ near the uniform prior $1/(n+1)$, thereby eliminating unobservable or irrelevant regions.

\subsubsection{Fast Scene Refinement with New Images}
Our reconstruction begins with few initial views with predefined camera pose. Given $m$ views, we run optimization for $100m$ iterations and then proceed with the \ac{NBV} selection, as described in Section 3.B, based on the current reconstruction. To enable faster optimization during the early stages, we set the spherical harmonic (SH) degree $d=0$ until the final view is added. Additionally, to avoid the influence of local minima or overfitting effects from previous reconstructions, we reinitialize the Gaussians with new random points at each step. The overall loss function is based on the standard 3DGS loss, with an additional term introduced for supervising the object vector associated with each Gaussian.

% 우리의 active reconstruction은 정해진 2~4개의 view로부터 3d를 reconstruction하는 것이다. view가 $m$개 있을 때 iteration은 ${k}_{iter}m$번으로 잡고 학습을 돌리고 그 때까지의 reconstruction을 이용하여 3.B에서 소개될 NBV를 진행한다. reconstruction 과정에 sparse-view를 위한 기존에 제안된 방법들을 사용하지는 않으며, 빠른 optimization을 위해 마지막 view가 추가되기 이전까지 sh degree d는 0으로 잡는다. 또한, 이전 reconstruction에서 발생했을 local minimum이나 overfitting의 효과를 무시하기 위해 gaussian들의 initialization 또한 다시 random points로 한다. loss는 기존 3dgs의 loss에 object vector를 위한 loss를 추가한다.

{\small
\begin{equation}
L_{overall} = (1-\lambda) L_{1} + \lambda L_{SSIM} + \lambda_{obj}L_{obj}
\end{equation}
}

% While standard \ac{3DGS} typically requires around 30,000 iterations, integrating segmentation significantly reduces this requirement. Once the designated number of views has been collected, the scene can be reconstructed in only 3,000 iterations, despite raising the SH degree back to 3. Even when considering all iterations performed before the full view set is collected, the total number of iterations remains under 10,000, which is one-third of that in the original setup.

Once the designated number of views has been collected, the final reconstruction can be achieved in 3,000 iterations, despite raising the SH degree back to 3 because we integrate object segmentation into the optimization. Even when considering all iterations performed before the full view set is acquired, the total iterations remain under 10,000, compared to 30,000 iterations in the standard \ac{3DGS} setup.

% segmentation의 또다른 장점은 빠른 optimization이다. 기존의 3dgs가 아무 설정 없이 30000번의 iteration을 필요로 한 것과 달리, segmentation을 포함하면 정해진 개수의 뷰를 모은 뒤 sh degree를 다시 3으로 올린 뒤에도 기존의 1/10인 3000번의 iteration 만으로 scene을 빠르게 reconstruct할 수 있다. 이는 단순히 iteration이 적은 것에 더해, background pruning을 통해 gaussian의 수가 준 것에 대한 효과도 있다. 뷰를 전부 모으기 전의 iteration을 전부 포함해도 전체 iteration은 설정에 따라 다르지만 10000번 이하로 reconstruction을 끝낼 수 있다.

\begin{figure}[!t]
  \centering
  \includegraphics[width=1\linewidth]{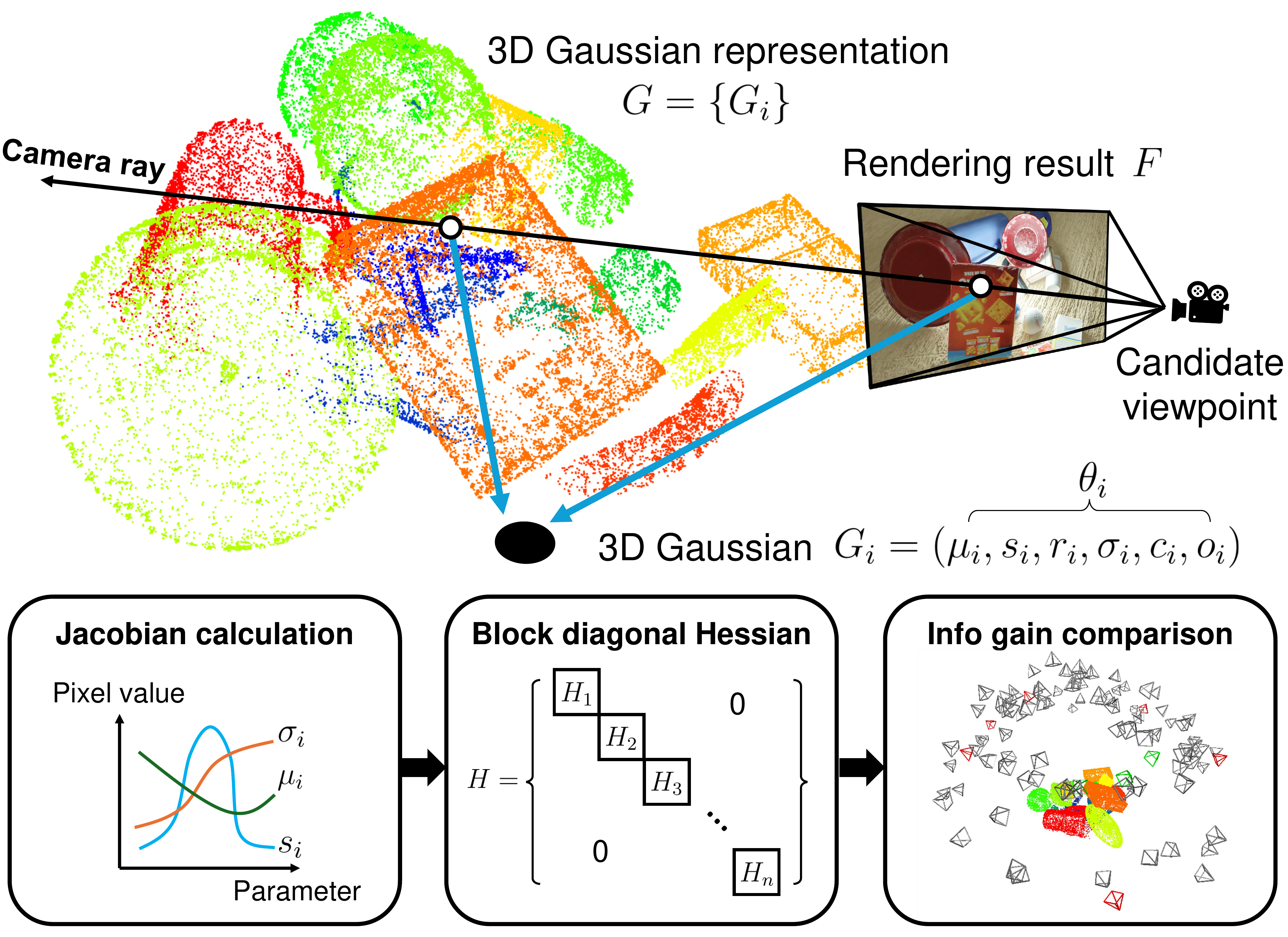}
  % \fbox{\rule{0pt}{0.5\columnwidth} \rule{0.9\columnwidth}{0pt}}
  \caption{Illustration of the overall NBV system. Given an object-segmented 3D Gaussian map $G$ and candidate camera pose, the CUDA rasterizer computes the Jacobian of the rendering with respect to Gaussian parameters. Correlations among parameters of each Gaussian are captured in a block diagonal Hessian, which is then used to estimate information gain. By comparing gains across candidates, the most informative view is selected, without requiring GT images of candidate views.}
  % 전반적인 NBV system에 대한 illustration. 3D Gaussian map G와 candidate camera pose p가 주어질 때, 먼저 gaussian들의 파라미터들에 대한 해당 candidate에서의 rendering의 jacobian을 CUDA rasterizer에서 계산한다. 같은 gaussian에서의 파라미터들간의 correlation이 고려되어 Block diagonal Hessian을 구할 수 있는데, candidate view의 유무에 따른 information gain을 비교하여 가장 informative한 view를 선택한다. 이 과정에서 candidate view의 GT는 사용하지 않는다.
  \label{fig:uncertainty}
  \vspace{-5mm}
\end{figure}

\subsection{Next Best View Policy with Uncertainty Quantization}
\subsubsection{Information Gain of the New Image}
% To quantify information gain from adding a view, we interpret how much each Gaussian’s parameters can vary without changing the rendering results on the training views as epistemic uncertainty. 
Information gain $IG(T,c)$ with training view set $T$ and candidate $c$ can be calculated by comparing quantified uncertainty value with and without the new image. Candidate which maximizes the information gain among the candidate view set $C$ includes to the training view set:

{\small
\begin{equation}
IG(T,c) = f(T \cup c) - f(T), \quad c=\operatorname*{argmax}_{c \in C} IG(T,c).
\end{equation}
}

Following POp-GS \cite{wilson2025pop}, we assume i.i.d. Gaussian pixel noise and adopt a maximum-likelihood formulation as shown in \figref{fig:uncertainty}. 
Maximizing the likelihood is equivalent to minimizing the least-squares objective $\frac{1}{2}||e||^{2}$ for the \ac{3DGS} model $h$, where $e$ denotes the per-pixel residual between the observation and the rendering.
% Under this assumption, the negative log-likelihood reduces to the least-squares objective $\frac{1}{2}||e||^{2}$ for the \ac{3DGS} model $h$, where $e$ is the per-pixel residual between the observation and the rendering.
% 이 문장 뭔 소리지
% Following POp-GS \cite{wilson2025pop}, we adopt a maximum-likelihood formulation with IID Gaussian pixel noise, where the negative log-likelihood reduces to a least-squares objective $\frac{1}{2}||e||^{2}$ for the \ac{3DGS} model $h$ as illustrated in \figref{fig:uncertainty}. 
Linearizing $h(\theta)$ around the current estimate yields the Gauss–Newton approximation with the Hessian $H \approx J^\top J$, where $J$ is the Jacobian of the rendering with respect to Gaussian parameters. We use $H^{-1}$ as a proxy for epistemic uncertainty and score candidate views by their expected reduction of this covariance while preserving fidelity on the training views.

Several choices exist for $f$ such as the trace-based $trace(H^{-1})^{-1}$ or the determinant $\det(H)$ of the Hessian, but we adopt the log-determinant $\log(\det(H))$ for numerical stability and computational convenience. Given a view set $V$, the matrix $H$ is computed as the sum of $J^\top J$ over all views in $V$:

{\small
\begin{equation}
f(V) = \log(\det(H_V)) = \log( \det(\sum_{v \in V}J_v^\top J_v)).
\end{equation}
}

Note that this process does not require ground-truth images; it relies solely on camera poses, making it applicable to candidate views as well.

% f는 trace나 determinant 등 다양한 값을 사용 가능하나, 우리는 log(det(H))를 사용한다. 주어진 view set V에 대해 H는 V에 대한 view 각각에 대한 $J^TJ$의 합으로 정의된다. Note that 이 과정에서는 view의 gt image를 필요로 하지 않고 오직 camera pose만을 사용하므로 candidate에 대해서도 정상적으로 사용할 수 있다.

For $n$ Gaussians each with $l$ parameters, the full Information matrix $H_V$ has a dimensionality of $nl \times nl$, which results in a substantial computational burden when computing its determinant. To alleviate this, we consider only the intra-Gaussian correlations among parameters, approximating $H_V$ as a block diagonal matrix. This allows us to compute the determinant as the product of per-Gaussian blocks, reducing the computational complexity from $O(l^3n^3)$ to $O(l^3n)$. This block-diagonal approximation is implemented by modifying the original CUDA rasterizer in \ac{3DGS}.

% 한 개당 l개의 parameter를 가지고 있는 gaussian이 n개라고 할 때 H_V는 $R^{nl x nl}$의 dimension을 가지고, 이는 detminant을 계산할 때 굉장한 computational cost를 필요로 한다. 이를 줄이기 위해 우리는 같은 gaussian의 parameter에 대한 correlation만을 고려하고 block diagonal로 Hessian matrix의 determinant를 계산하여 계산량을 O(l^3n^3)에서 O(l^3n)으로 줄였다. 이 과정은 기존 \ac{3DGS}의 cuda rasterizer를 변형하여 구현되었다.

\definecolor{mypurple}{HTML}{7600bc}
\definecolor{mypink}{HTML}{ff69b4}
\begin{figure}[!t]
  \centering
  \includegraphics[width=1\linewidth]{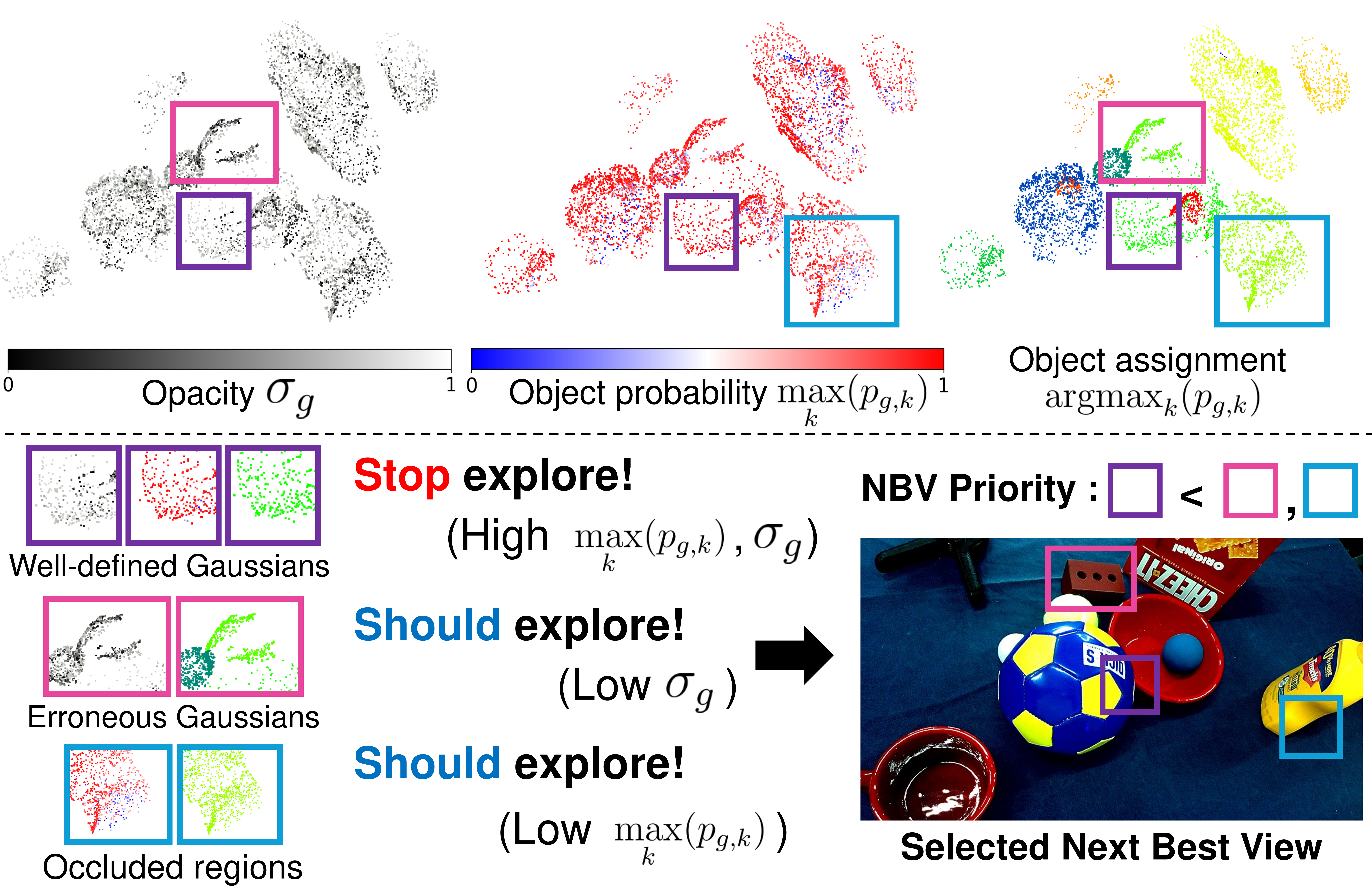}
  % \fbox{\rule{0pt}{0.3\columnwidth} \rule{0.9\columnwidth}{0pt}}
  \caption{Illustration of object confidence scores for exploration. The top row shows the 3D representation of a captured scene trained with 6 views, visualized as Gaussian opacity, object probability, and object index. Rectangles of the same color highlight different cases: \textcolor{mypurple}{purple} indicates well-optimized regions that require no further observation, while \textcolor{mypink}{pink} and \textcolor{blue}{blue} denote areas either unseen or poorly localized, thus requiring exploration. We leverage the intuition that regions needing additional views exhibit low opacity and low object probability, and incorporate this insight into our NBV formulation.}
  % Exploration을 위한 object confidence score에 대한 illustration이다. 윗줄의 세 이미지는 6장을 이용해 학습한 captured scene의 3d representation으로, 왼쪽부터 gaussian의 opacity, object probability, object idx이다. 같은 색깔의 직사각형들을 보면 보라색은 optimize가 잘 되었기 때문에 더 볼 필요 없고, 분홍색과 파란색은 아직 못 봤거나 제대로 위치하지 않기 때문에 exploration이 필요한 상황이다. 우리는 뷰가 더 필요한 경우 opacity와 object probability가 낮다는 직관을 이용하여 nbv에 이를 수식적으로 적용하였다.
  \label{fig:object confidence}
  \vspace{-5mm}
\end{figure}

\subsubsection{Confidence-weighted Information Gain for Exploration}
Gaussians corresponding to regions that require further observation through additional views tend to have low contributions in the training views. As a result, they generally exhibit small gradients, low opacity $\sigma_i$, and uniform object probability $p_{i}$ that are not strongly biased toward a single object. Consequently, in Hessian-based \ac{NBV} formulations, the Jacobians of well-observed primitives with low uncertainty still dominate the overall information, leading to a misalignment between information gain and true exploration needs. To address this issue, we propose a confidence-weighted information matrix that incorporates confidence of each Gaussian, prioritizing under-observed regions.

% 우리가 추가적인 view를 통해 봐야할 region의 gaussian들은 training view에서의 기여도가 적기 때문에 일반적으로 작은 gradient와 opacity, 완전히 치우쳐져 있지 않은 object vector를 가지고 있다. 따라서 Hessian matrix에 기반한 NBV는 observation이 많아 uncertainty가 작은 primitive의 jacobian이 여전히 전체 information에 dominant한 영향을 끼친다는 문제를 가지고 있다. 이런 misalignment을 해소하기 위해 우리는 gaussian 별로의 confidence를 고려한 confidence-weighted information matrix를 제안한다.

{\small
\begin{equation}
c_g=\max_k(p_{g,k})^{-\alpha_{obj}}{\sigma_g}^{-\alpha_{opa}}
\end{equation}
\begin{equation}
C_g=\text{diag}(c_g,...c_g) \in R^{l \times l}
\end{equation}
}

We define confidence without introducing any additional components, instead leveraging the previously discussed insights on opacity and the object vector. For example, as illustrated in \figref{fig:object confidence}, the well-observed part of the surface typically exhibits both high opacity and a maximum object probability close to 1. On the other hand, Gaussians that are poorly fitted or occluded tend to exhibit low opacity and object probability. To encourage more exploration of low-confidence Gaussians and reduce focus on those already well observed, we scale each Jacobian by a factor inversely proportional to both the opacity and the object probability during information gain computation:

{\small
\begin{equation}
% H=J^\top C J, \quad C = \text{diag}(C_g).
% H\overset{\Delta}{=}J^\top C J, \quad C = \text{diag}(C_g).
H \coloneq J^\top C J, \quad C = \text{diag}(C_g).
\end{equation}
}

% 우리는 confidence를 구하기 위해 뭔가를 추가하지 않고, 앞에서 언급한 opacity와 object vector에 관한 insight를 이용하여 confidence를 정의한다. 예를 들어, observation이 많은 물체의 위쪽 표면의 경우 opacity와 object vector의 max 값이 1에 가깝다. 이렇게 높은 confidence의 gaussian들은 조금 덜 보고, 그렇지 않은 gaussian은 더 보게 하기 위해 두 값에 반비례하는 값을 jacobian에 곱하여 information gain을 측정했다.

\subsubsection{Information Gain from Multiple Outputs}
A single 3DGS map can render RGB images, depth maps, and object masks from the same Gaussians but with different parameter subsets. All outputs use position, scale, rotation, and opacity, while RGB additionally uses SH coefficients, and masks use the object vector. Although information gain can be aggregated across outputs, scale differences across parameters and outputs make it difficult to balance their contributions.

% 3dgs를 통해 우리는 한 gaussian map에서 rgb image, depth, object mask를 얻을 수 있고, 그 때마다 사용되는 parameter가 다르다. specifically, 셋 모두에 position, scale, rotation, opacity가 사용되고 rgb image와 object mask에서는 각각 sh parameter와 object vector가 추가로 사용된다. Next Best Sense에서는 information gain을 계산할 때 이 output들 모두에 대한 값을 더해서 NBV를 진행했지만, 이 경우엔 output이나 parameter 간의 scale issue로 인해 발생하는 imbalance를 balancing하기 어렵다는 것이다.

We address this issue with the following insight: for a well-optimized Gaussian map, treating any training view as a candidate should yield negligible—ideally zero—additional information gain. In practice, however, Hessian-based information gain does not satisfy this property, since the Jacobian $J$ is not zero for training views. Hessian matrix $H'=H+J^\top J$ contains more information than $H$, thereby inflating the gain.

{\small
\begin{equation}
\tilde{IG}_f(T,c) = \frac{IG_f(T,c)}{\frac{1}{|T|} \sum_{t \in T} IG_f(T,t)} \quad f \in \left\{ rgb, d, o \right\}
\end{equation}
}

To compensate, we normalize each output’s information gain by the mean information gain over the training views, which calibrates modality scales so that a typical training view has unit gain. This per-output normalization mitigates imbalance and allows RGB, depth, and mask contributions to be integrated more fairly and effectively in \ac{NBV}:

{\small
\begin{equation}
\tilde{IG}(T,c) = \tilde{IG}_{rgb}(T,c) + \beta_{d}\tilde{IG}_{d}(T,c) + \beta_{o}\tilde{IG}_{o}(T,c)
\end{equation}
}

% 이를 위해 우리는 하나의 insight를 활용하였는데, training view에 대해 잘 optimize된 3dgs gaussian map은 training view를 candidate처럼 생각할 때 추가적인 information gain이 매우 작고 이론적으로 0이어야 한다는 것이다. 그러나, Hessian matrix를 기반으로 한 information gain은 계산상으로 training view에 대해 J가 0이 아니기 때문에 H' = H + J^TJ에서 H'이 H보다 많은 information을 가지고 있는 것이 된다. 따라서 우리는 이를 output 간 scaling에 사용하여 같은 candidate view의 multiple output의 information gain에 대해 각각을 training view들의 information gain의 평균으로 나눈다. 식으로 표현하면 아래와 같다.

\begin{figure*}[!t]
  \centering
  % \fbox{\rule{0pt}{0.2\textwidth} \rule{0.9\textwidth}{0pt}}
  \includegraphics[width=\linewidth]{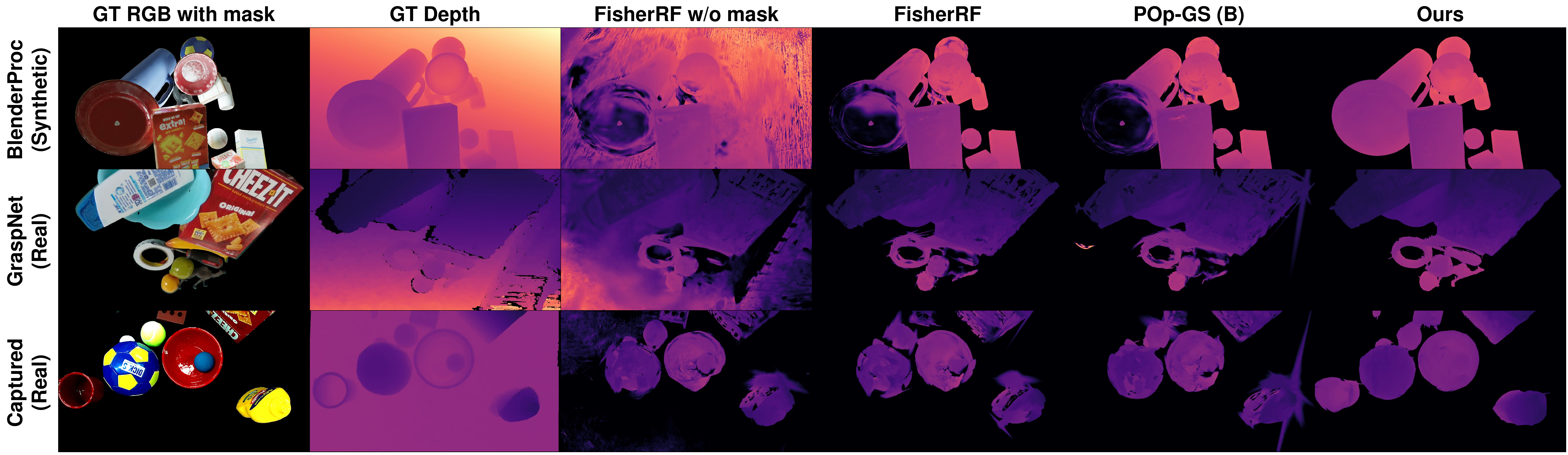}
  \caption{Qualitative results of NBV system and 3D reconstruction for whole scene. Looking at the dish or the snack box, we can see that in the baseline, since training was performed only with RGB without instance information, textures and text appear in the depth results. In contrast, with our method such artifacts are absent, and the views selected through our NBV capture scene information from all directions, enabling more accurate overall depth reconstruction.}
  % 첫 Row의 접시나 두 번째 과자 박스를 보면 baseline에서는 semantic 정보를 주지 않고 rgb로만 학습했기 때문에 texture나 글씨가 depth 결과에 드러나는 것을 볼 수 있다. 반면 우리는 이런 모습이 보이지 않고 우리의 NBV를 통해 뽑힌 뷰가 scene의 information을 전방위로 적절히 뽑아주기에 전반적인 depth도 정확하게 복원한다. 
  \label{fig:fig5}
  \vspace{-1mm}
\end{figure*}

\begin{table*}[ht]
\centering

\caption{Results of \ac{NBV} evaluation for whole scene reconstruction in synthetic and GraspNet dataset. Best results are highlighted in \textbf{bold}, and second-best results are \underline{underlined}. The “All” row, which uses all candidate views, is excluded from this comparison.}
% 모든 candidate를 사용한 "All"을 제외하고 Best는 Bold, second best는 밑줄을 쳤다.
\vspace{-2mm}
\label{tab:main}

\resizebox{\textwidth}{!}{
\renewcommand{\arraystretch}{1.2}
{\large
\begin{tabular}{cc|cccccccc|cccccccc}
\hline \toprule
\multicolumn{2}{c|}{\multirow{3}{*}{\textbf{Methods}}} &
  \multicolumn{8}{c|}{\textbf{BlenderProc Synthetic Dataset (10 Scenes)}} &
  \multicolumn{8}{c}{\textbf{GraspNet Real Dataset (10 Scenes)}} \\ \cline{3-18} 
\multicolumn{2}{c|}{} &
  \multicolumn{4}{c|}{\textbf{W/o Obj mask}} &
  \multicolumn{4}{c|}{\textbf{With Obj mask}} &
  \multicolumn{4}{c|}{\textbf{W/o Obj mask}} &
  \multicolumn{4}{c}{\textbf{With Obj mask}} \\ \cline{3-18} 
\multicolumn{2}{c|}{} &
  PSNR $\uparrow$ &
  SSIM $\uparrow$ &
  LPIPS $\downarrow$ &
  \multicolumn{1}{c|}{D-MAE $\downarrow$} &
  PSNR $\uparrow$ &
  SSIM $\uparrow$ &
  LPIPS $\downarrow$ &
  D-MAE $\downarrow$ &
  PSNR $\uparrow$ &
  SSIM $\uparrow$ &
  LPIPS $\downarrow$ &
  \multicolumn{1}{c|}{D-MAE $\downarrow$} &
  PSNR $\uparrow$ &
  SSIM $\uparrow$ &
  LPIPS $\downarrow$ &
  D-MAE $\downarrow$ \\ 
  \hline \midrule
  % \hline  
% \multicolumn{1}{c|}{No Selection} &
%   All &
  \multicolumn{2}{c|}{All} &
  23.87 &
  0.7756 &
  0.3249 &
  \multicolumn{1}{c|}{0.0390} &
  29.95 &
  0.9646 &
  0.0601 &
  0.0552 &
  22.49 &
  0.8035 &
  0.3885 &
  \multicolumn{1}{c|}{0.0645} &
  23.15 &
  0.8759 &
  0.1590 &
  0.0664 \\ 
  % \hline
  \midrule
% \multicolumn{1}{c|}{\multirow{3}{*}{\begin{tabular}[c]{@{}c@{}}Heuristic \\ Selection\end{tabular}}} &
  % Spiral &
  \multicolumn{2}{c|}{Spiral} &
  17.75 &
  0.5726 &
  0.4100 &
  \multicolumn{1}{c|}{\underline{0.0630}} &
  25.69 &
  0.8836 &
  0.1047 &
  0.0694 &
  19.85 &
  0.7326 &
  0.4164 &
  \multicolumn{1}{c|}{0.0749} &
  \textbf{21.69} &
  \underline{0.8414} &
  \textbf{0.1605} &
  \underline{0.0739} \\
% \multicolumn{1}{c|}{} &
%   Random &
  \multicolumn{2}{c|}{Random} &
  17.95 &
  0.5568 &
  0.4347 &
  \multicolumn{1}{c|}{0.0701} &
  24.91 &
  0.9079 &
  0.0978 &
  0.0716 &
  18.55 &
  0.6990 &
  0.4427 &
  \multicolumn{1}{c|}{0.0801} &
  20.44 &
  0.7997 &
  0.1830 &
  0.0786 \\
% \multicolumn{1}{c|}{} &
%   FPS &
  \multicolumn{2}{c|}{FPS} &
  18.42 &
  0.5808 &
  0.4271 &
  \multicolumn{1}{c|}{0.0670} &
  24.47 &
  0.9026 &
  0.1020 &
  0.0733 &
  19.22 &
  0.7098 &
  0.4404 &
  \multicolumn{1}{c|}{0.0822} &
  21.25 &
  0.8346 &
  0.1706 &
  0.0786 \\ 
  % \hline
  \midrule
% \multicolumn{1}{c|}{\multirow{5}{*}{\begin{tabular}[c]{@{}c@{}}Info-based\\ Selection\end{tabular}}} &
%   FisherRF &
  \multicolumn{2}{c|}{FisherRF \cite{jiang2024fisherrf}} &
  18.64 &
  0.5753 &
  0.4335 &
  \multicolumn{1}{c|}{0.0703} &
  25.34 &
  0.9253 &
  0.0873 &
  0.0695 &
  18.82 &
  0.7046 &
  0.4358 &
  \multicolumn{1}{c|}{0.0805} &
  21.22 &
  0.8328 &
  0.1715 &
  0.0768 \\
% \multicolumn{1}{c|}{} &
%   POp-GS (T) &
  \multicolumn{2}{c|}{POp-T \cite{wilson2025pop}} &
  19.07 &
  0.6046 &
  0.4093 &
  \multicolumn{1}{c|}{0.0648} &
  26.54 &
  0.9366 &
  0.0783 &
  0.0651 &
  19.05 &
  0.7047 &
  0.4435 &
  \multicolumn{1}{c|}{0.0801} &
  21.10 &
  0.8310 &
  0.1733 &
  0.0768 \\
% \multicolumn{1}{c|}{} &
%   POp-GS (D) &
  \multicolumn{2}{c|}{POp-D \cite{wilson2025pop}} &
  19.04 &
  0.5973 &
  0.4139 &
  \multicolumn{1}{c|}{0.0675} &
  26.64 &
  0.9387 &
  \underline{0.0770} &
  0.0641 &
  18.99 &
  0.7046 &
  0.4427 &
  \multicolumn{1}{c|}{0.0801} &
  21.16 &
  0.8313 &
  0.1731 &
  0.0770 \\
% \multicolumn{1}{c|}{} &
%   POp-GS (B) &
  \multicolumn{2}{c|}{POp-B \cite{wilson2025pop}} &
  19.04 &
  0.5973 &
  0.4139 &
  \multicolumn{1}{c|}{0.0675} &
  \underline{26.69} &
  \underline{0.9393} &
  \textbf{0.0766} &
  0.0636 &
  18.99 &
  0.7043 &
  0.4413 &
  \multicolumn{1}{c|}{0.0803} &
  21.12 &
  0.8315 &
  0.1731 &
  0.0768 \\
% \multicolumn{1}{c|}{} &
%   \textbf{Ours} &
  \multicolumn{2}{c|}{\textbf{Ours}} &
  \textbf{-} &
  - &
  - &
  \multicolumn{1}{c|}{-} &
  \textbf{26.90} &
  \textbf{0.9408} &
  0.0773 &
  \textbf{0.0144} &
  - &
  - &
  - &
  \multicolumn{1}{c|}{-} &
  \underline{21.47} &
  \textbf{0.8430} &
  \underline{0.1653} &
  \textbf{0.0487} \\ 
  % \hline
  \bottomrule
\end{tabular}
}
}
\end{table*}

\begin{table*}[ht]
\centering

\caption{Results of \ac{NBV} evaluation for whole scene reconstruction in captured scenes. Best results are highlighted in \textbf{bold}, and second-best results are \underline{underlined}. The “All” row, which uses all candidate views, is excluded from this comparison.}
% N/A는 모든 실험에서 지나치게 많은 Gaussian으로 인해 memory ood가 된 경우이다.
\vspace{-2mm}
\label{tab:main2}

\resizebox{\textwidth}{!}{
\renewcommand{\arraystretch}{1.2}
{\large
\begin{tabular}{cc|cccccccc|cccccccc}
\hline
\toprule
\multicolumn{2}{c|}{\multirow{3}{*}{\textbf{Methods}}} &
  \multicolumn{8}{c|}{\textbf{Captured Scene - Grounded SAM2 Mask (2 scenes)}} &
  \multicolumn{8}{c}{\textbf{Captured Scene - Manual SAM Mask (2 scenes)}} \\ \cline{3-18} 
\multicolumn{2}{c|}{} &
  \multicolumn{4}{c|}{\textbf{W/o Obj mask}} &
  \multicolumn{4}{c|}{\textbf{With Obj mask}} &
  \multicolumn{4}{c|}{\textbf{W/o Obj mask}} &
  \multicolumn{4}{c}{\textbf{With Obj mask}} \\ \cline{3-18} 
\multicolumn{2}{c|}{} &
  PSNR $\uparrow$ &
  SSIM$\uparrow$ &
  LPIPS $\downarrow$ &
  \multicolumn{1}{c|}{D-MAE $\downarrow$} &
  PSNR $\uparrow$ &
  SSIM$\uparrow$ &
  LPIPS $\downarrow$ &
  D-MAE $\downarrow$ &
  PSNR $\uparrow$ &
  SSIM$\uparrow$ &
  LPIPS $\downarrow$ &
  \multicolumn{1}{c|}{D-MAE $\downarrow$} &
  PSNR $\uparrow$ &
  SSIM$\uparrow$ &
  LPIPS $\downarrow$ &
  D-MAE $\downarrow$ \\
  % \hline
  \hline \midrule
\multicolumn{2}{c|}{All} &
  17.04 &
  0.6439 &
  0.5401 &
  \multicolumn{1}{c|}{0.0985} &
  17.23 &
  0.8333 &
  0.2053 &
  \multicolumn{1}{c|}{0.1169} &
  17.85 &
  0.7049 &
  0.5145 &
  \multicolumn{1}{c|}{0.0561} &
  18.27 &
  0.7839 &
  0.2936 &
  0.0494  \\ 
  % \hline
  \midrule
% \multicolumn{1}{c|}{\multirow{3}{*}{\begin{tabular}[c]{@{}c@{}}Heu\\ristic\end{tabular}}} &
  % Spiral &
\multicolumn{2}{c|}{Spiral} &
  15.51 &
  0.5706 &
  0.5566 &
  \multicolumn{1}{c|}{\underline{0.1066}} &
  \underline{16.66} &
  \underline{0.8072} &
  \underline{0.2152} &
  \multicolumn{1}{c|}{0.1128} &
  14.56 &
  0.5424 &
  0.5714 &
  \multicolumn{1}{c|}{0.0845} &
  15.49 &
  0.6554 &
  0.3511 &
  0.0792 \\
% \multicolumn{1}{c|}{} &
%   Random &
\multicolumn{2}{c|}{Random} &
  14.42 &
  0.5318 &
  0.5759 &
  \multicolumn{1}{c|}{0.1178} &
  16.22 &
  0.7668 &
  0.2315 &
  \multicolumn{1}{c|}{0.1228} &
  15.11 &
  0.5807 &
  0.5578 &
  \multicolumn{1}{c|}{0.0841} &
  15.57 &
  0.6699 &
  0.3397 &
  0.0779 \\
% \multicolumn{1}{c|}{} &
%   FPS &
\multicolumn{2}{c|}{FPS} &
  14.67 &
  0.5404 &
  0.5736 &
  \multicolumn{1}{c|}{0.1239} &
  16.37 &
  0.7787 &
  0.2271 &
  \multicolumn{1}{c|}{0.1336} &
  14.23 &
  0.5471 &
  0.5775 &
  \multicolumn{1}{c|}{0.0811} &
  15.26 &
  0.6554 &
  0.3543 &
  \underline{0.0637} \\ 
  % \hline
  \midrule
% \multicolumn{1}{c|}{\multirow{5}{*}{\begin{tabular}[c]{@{}c@{}}Info\\based\end{tabular}}} &
%   FisherRF &
\multicolumn{2}{c|}{FisherRF \cite{jiang2024fisherrf}} &
  14.28 &
  0.5127 &
  0.5852 &
  \multicolumn{1}{c|}{0.1283} &
  15.60 &
  0.7677 &
  0.2405 &
  \multicolumn{1}{c|}{0.1135} &
  14.91 &
  0.5739 &
  0.5646 &
  \multicolumn{1}{c|}{0.0857} &
  15.34 &
  0.6697 &
  0.3472 &
  0.0761 \\
% \multicolumn{1}{c|}{} &
%   POp-GS (T) &
\multicolumn{2}{c|}{POp-T \cite{wilson2025pop}} &
  14.70 &
  0.5255 &
  0.5728 &
  \multicolumn{1}{c|}{0.1265} &
  16.16 &
  0.7883 &
  0.2271 &
  \multicolumn{1}{c|}{0.1291} &
  14.96 &
  0.5630 &
  0.5638 &
  \multicolumn{1}{c|}{0.0803} &
  15.44 &
  0.6837 &
  0.3412 &
  0.0766 \\
% \multicolumn{1}{c|}{} &
%   POp-GS (D) &
\multicolumn{2}{c|}{POp-D \cite{wilson2025pop}} &
  14.93 &
  0.5319 &
  0.5678 &
  \multicolumn{1}{c|}{0.1224} &
  16.28 &
  0.7894 &
  0.2264 &
  \multicolumn{1}{c|}{0.1188} &
  15.18 &
  0.5783 &
  0.5596 &
  \multicolumn{1}{c|}{0.0864} &
  \underline{15.65} &
  \underline{0.6927} &
  \underline{0.3350} &
  0.0773 \\
% \multicolumn{1}{c|}{} &
%   POp-GS (B) &
\multicolumn{2}{c|}{POp-B \cite{wilson2025pop}} &
  14.50 &
  0.5251 &
  0.5761 &
  \multicolumn{1}{c|}{0.1264} &
  16.29 &
  0.7902 &
  0.2263 &
  \multicolumn{1}{c|}{0.1214} &
  15.20 &
  0.5808 &
  0.5618 &
  \multicolumn{1}{c|}{0.0823} &
  15.58 &
  0.6895 &
  0.3385 &
  0.0747 \\
% \multicolumn{1}{c|}{} &
%   \textbf{Ours} &
\multicolumn{2}{c|}{\textbf{Ours}} &
  - &
  - &
  - &
  \multicolumn{1}{c|}{-} &
  \textbf{16.74} &
  \textbf{0.8148} &
  \textbf{0.2064} &
  \multicolumn{1}{c|}{\textbf{0.0527}} &
  - &
  - &
  - &
  \multicolumn{1}{c|}{-} &
  \textbf{16.71} &
  \textbf{0.7280} &
  \textbf{0.3102} &
  \textbf{0.0461} \\ 
  % \hline
  \bottomrule
\end{tabular}
}
}
\vspace{-4mm}
\end{table*}

\subsection{Object-centric Next Best View for Manipulation}

When a robot is tasked with performing a specific operation, it is often unnecessary to reconstruct the entire scene. If a particular object or region is of interest, prioritizing the reconstruction of that area is more practical and efficient. To enable this targeted reconstruction, we apply a simple modification to the previously defined confidence measure. Instead of using the maximum value of the object vector for each Gaussian, we directly use the value corresponding to the target object’s for confidence:

{\small
\begin{equation}
c_{g,obj}=\begin{cases} 
		p_{g,obj}^{-\alpha_{obj}} {\sigma_g}^{-\alpha_{opa}} & \text{if } \text{argmax}_{k}(p_{g,k}) = obj \\ 
         0 & \text{else }
     \end{cases},
\end{equation}
}
and this allows the system to focus on the Gaussians most relevant to the specified object.

% 로봇이 특정 task를 수행할 때 전체 scene을 굳이 전부 reconstruction 할 필요는 없다. 만약 특정 object나 영역에 관심이 있다면, 그 물체 주변을 우선적으로 reconstruction 하는 것이 필요하다. 우리는 집중하고 싶은 물체가 주어졌을 때 이를 가능하게 하기 위해 이전의 confidence에 간단한 트릭을 넣었다. 그것은 바로 object vector의 max를 사용하는 것이 아닌, 그저 그 object에 대한 element를 사용하는 것이다.

After performing reconstruction using the views selected through NBV, we obtain depth images from all training views for actual manipulation. In cluttered scenes, feeding a single bird’s-eye view depth image into the grasping model is often suboptimal. Since the selected views are expected to contain high-information content, it is more effective to extract candidate grasping points from the depth maps of these views and select the most confident one among them.

% 이렇게 nbv를 통해 얻은 view로 reconstruction을 한 뒤, 실제 manipulation을 하기 위해 모든 training view에서 depth를 얻는다. cluttered scene에서는 보통의 방식대로 단순히 bird-eye view에서의 depth를 모델에 넣는 것은 권장되지 않는다. 우리가 얻은 view는 information이 많을 것으로 예상되는 view들이었기 때문에, 그 view들의 depth로부터 얻을 수 있는 여러 grasping point 중 가장 좋은 것을 선택하는 것이 좋다. 

\section{experiment}
\label{sec:experiment}

\begin{figure*}[!t]
  \centering
  \includegraphics[width=\linewidth]{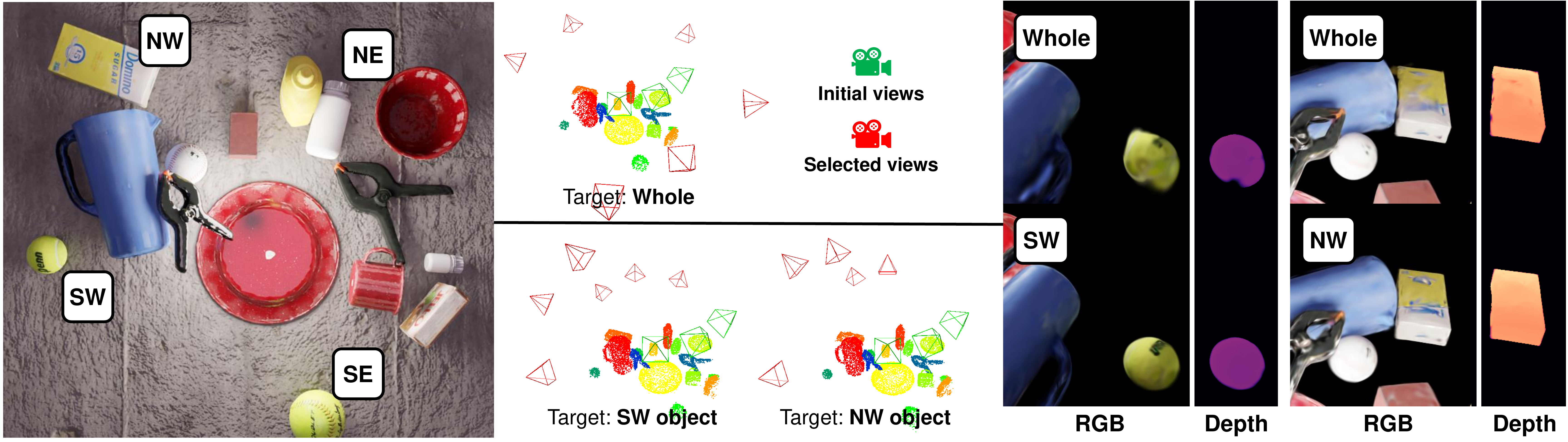}
  % \fbox{\rule{0pt}{0.2\textwidth} \rule{0.9\textwidth}{0pt}}
  \caption{\textbf{Left}: Cluttered scene with objects far from the center, labeled using cardinal directions (NW, NE, SW, SE). \textbf{Center}: Views selected by NBV (red camera frustum) targeting the entire scene versus object-centric NBV focusing on the SW and NW objects. While targeting the entire scene yields a near-uniform distribution of viewpoints, specifying a particular object (the SW or NW object) results in viewpoints clustering on the western side, from which the target object is best observed. \textbf{Right}: Reconstructions trained with these selected views yield improved RGB and depth renderings compared with whole scene NBV.}
  % 왼쪽은 cluttered 씬에서 중심에서 떨어져 있는 물체들을 표시한 것으로, 방위를 이용하여 NW, NE, SW, SE로 나타낸 것이다. 가운데는 전체 scene을 목표로 NBV를 했을 때 선택되는 뷰 (빨간색 카메라 frustum)과 SW, NW 물체에 집중된 NBV를 했을 때 선택되는 뷰를 visualize한 것으로, 해당 물체를 주로 보는 뷰가 집중적으로 선택되는 것을 볼 수 있다. 이렇게 선택된 뷰로 학습한 reconstruction은 오른쪽에서 보는 것처럼 전체 scene을 목표로 했을 때보다 더 좋은 rgb 및 depth rendering 결과를 보여준다.
  \label{fig:object centric recon}
  \vspace{-5mm}
\end{figure*}

In this section, we evaluate the effectiveness of our view selection for both reconstruction and manipulation. First, we assess how closely the reconstructed scene obtained from the selected views matches the ground-truth geometry. Next, we evaluate the quality of reconstruction for a specific object of interest to determine whether the selected views effectively focus on that target. Finally, we apply our method to real-world robotic experiments.

% 이 섹션에서는 앞에서 소개한 method에 따라 선택된 view가 reconstruction과 manipulation에 얼마나 효과적인지를 확인할 것이다. 우선 뷰를 고른 뒤 같은 방법으로 scene을 reconstruction 했을 때 fitting된 모델이 얼마나 실제와 비슷한지 비교할 것이다. 그 뒤, 집중하고 싶은 물체가 있을 때 고른 view들이 그 물체 한정으로 얼마나 잘 reconstruction하는 지 확인할 것이다. 마지막으로, 이 방법을 실제 robot experiment에 적용하여 잘 적용되는지 확인할 것이다.

\subsection{Next Best View for Whole Scene Reconstruction}
\subsubsection{Datasets and Metrics}

We evaluate our method on both real and synthetic datasets. For the real-world data, we use 10 scenes from the GraspNet \cite{fang2020graspnet} training set and 4 robot-captured scenes of BOP objects. GraspNet provides ground-truth object masks, while the captured scenes we use pseudo ground-truth masks generated with manual SAM or Grounded SAM2 \cite{ren2024grounded} with text prompts. For the synthetic data, we construct 10 scenes using BlenderProc \cite{denninger2019blenderproc} with BOP objects introducing heavy occlusions, leaning, and stacking among objects. Candidate views are generated on a sphere centered at the scene centroid, by combining randomly sampled viewpoints with spiral-arranged viewpoints that are evenly spaced in longitude at the same latitude.

% For the synthetic data, we construct 10 scenes using BlenderProc \cite{denninger2019blenderproc} with BOP objects. These synthetic scenes are specifically designed to challenge standard top-down perception by introducing heavy occlusions, leaning, and stacking among objects—conditions under which grasping points cannot be reliably determined from a single bird’s-eye view.

% 우리는 우리의 purposed NBV system을 real과 synthetic 양쪽에서 확인하였다. real dataset은 GraspNet의 학습 데이터셋의 한 scene과 MipNeRF360의 kitchen scene을 사용하였다. Object mask에 관하여 GraspNet은 gt object mask를 제공하여 그것을 사용하였고, MipNeRF360은 Grounded SAM2 를 이용하여 pseudo gt object mask를 만들어서 사용하였다. synthetic dataset은 BlenderProc을 이용하여 BOP dataset의 물체들로 scene 2개를 만들어서 사용하였다. 특히나 synthetic dataset은 물체들이 서로 크게 가리거나 기대거나 위에 올려져있는 등 Bird eye view 만으로 grasping point를 제대로 알기 어렵도록 만들어졌다.

% The primary evaluation metrics for reconstruction are PSNR, SSIM \cite{wang2004image}, and LPIPS \cite{zhang2018unreasonable}, computed from the rendered RGB images. To further assess geometric accuracy, we also compare the reconstructed depth MAE with the ground-truth depth. Since our method prunes Gaussians corresponding to background regions, depth comparisons are performed only on pixels that fall within the objects. All results are averaged over three runs.
The primary evaluation metrics for reconstruction are PSNR, SSIM, and LPIPS, computed from the rendered RGB images. To further assess geometric accuracy, we also compare the reconstructed depth MAE with the ground-truth depth. Since our method prunes Gaussians corresponding to background regions, depth comparisons are performed only on pixels that fall within the objects. All results are averaged over three runs.

% Reconstruction에 대한 기본적인 metric은 rgb rendering에 대한 psnr, ssim, lpips이다. 추가로, geometry에 집중하기 위해 gt depth에 대한 비교도 하여 MAE와 RMSE를 비교하였다. 우리의 방법은 background에 해당하는 gaussian들이 prune 되기 때문에 depth에 대한 비교는 gt object mask에 해당하는 부분에 대해서만 하였다.

\subsubsection{Baseline and Implementation Details}
For heuristic baselines, we use random sampling, evenly spaced spiral trajectories, and farthest point sampling. For information-based \ac{NBV}, we evaluate FisherRF \cite{jiang2024fisherrf} and the T-/D-optimality criteria from POp-GS \cite{wilson2025pop}, including a block-diagonal variant (B) that accounts for parameter correlations. (POp-T/D/B in table) Each baseline is tested with and without segmentation masks, which are used only for background removal without incorporating any additional features.

% view를 선택하는 가장 기본적인 방법은 random과 spiral 모양으로 일정한 간격으로 찍는 것이 있다. 추가로, 우리는 candidate 중 camera들 간의 거리가 가장 먼 farthest point들을 추가하였다. 추가로 next best view를 고르기 위한 방법 중 FisherRF와 Pop-gs의 T-optimality, D-optimality를 선택했다. baseline들의 경우 segmentation mask를 포함한 것과 포함하지 않은 것을 모두 하였다. 

Reconstruction follows the original \ac{3DGS} pipeline for initialization, rasterization, and densification, except for the object vector supervision, which is unique to our method. We set $\lambda_{Dice} = 0.5$, $\lambda_{obj}=0.1$, $\delta_{obj}=0.1$ for the reconstruction loss. For \ac{NBV}, confidence weights are $\alpha_{obj}=0.3$ and $\alpha_{opa}=0.3$. These values are selected through grid search and fixed for all scenes in our experiments. Information gain weights are $\beta_{d}=10$ and $\beta_{o}=1$ following \cite{strong2024next}.

% reconstruction을 할 때에는 object vector에 관한 부분을 제외하면 기존의 \ac{3dgs}와 hyperparameter의 값과 iteration을 제외한 initialization, rasterization, densification 등의 과정은 동일하게 하였다. reconstruction 과정에서 추가적인 hyperparameter에 대해서 lambda_Dice는 0.5, lambda_obj는 0.1이었고, NBV 과정에서 alpha_obj는 3, alpha_opa는 1로 하였다. scaling을 제외한 render, depth, mask information gain의 weight는 동일하게 1이다.

\subsubsection{Evaluation and Analysis}
\tabref{tab:main} and \tabref{tab:main2} show the results of reconstructing the entire scene by adding 6 views to an initial 4 views. In both the synthetic and real datasets, our method outperforms other information-based approaches. While methods mostly benefit from using object masks to remove the background, our approach achieves a particularly significant reduction in depth error thanks to the incorporation and optimization of object features, which effectively encodes information from object mask. Moreover, even for the captured scenes in \tabref{tab:main2}, where pseudo ground-truth object masks are used instead of actual ground-truth, our method continues to demonstrate strong performance.

% table 1,2는 한 쪽에 치우쳐진 4개의 initial view가 주어질 때 6개의 view를 추가하여 전체 scene을 reconstruction 했을 때의 결과이다. synthetic dataset과 real dataset 양쪽에서 우리의 방법이 다른 information-based method 보다 좋은 것을 확인할 수 있다. 모든 method에서 object mask를 이용하여 background를 없애는 것이 성능 향상을 이끌어냈지만, 특히 우리의 방법에서 one-hot object vector의 추가와 optimization으로 인해 semantic 정보가 담김으로서 depth error가 크게 줄어들었음을 확인할 수 있다. 또한, ground-truth가 아닌 grounded sam2로 얻은 pseudo ground-truth를 사용한 table 2의 captured scene들에 대해서도 여전히 좋은 성능을 보여준다.

The “All” row, which uses all candidate views for training, can be regarded as the upper bound for reconstruction within the scene. The spiral selection strategy intuitively produces a set of views that provides a well-balanced coverage of the scene and indeed outperforms other heuristic selection methods and information-based baselines in terms of reconstruction quality. However, because the spiral pattern does not account for scene-specific characteristics such as objects located deep inside, so it fails to adapt to individual scene layouts. This limitation becomes more pronounced in the synthetic dataset, which is designed to be more dynamic, compared to the simpler GraspNet dataset where training and test view distributions are more similar.

% 100장의 모든 candidate를 사용하여 학습한 'All' row는 해당 신에서 각 metric의 상한을 보여준다고 할 수 있다. spiral하게 view를 선택한 뷰는 직관적으로도 scene을 전반적으로 잘 표현하는 set이고, 실제로 다른 heuristic selection method들과 information 기반의 baseline들보다도 좋은 reconstruction 결과를 보여준다. 그러나 spiral view는 안쪽에 있는 물체 등은 고려하지 못하기 때문에 scene-specific하지 않고, 이는 scene이 단순하고 training view와 test view의 분포가 비슷한 graspnet dataset보다 더 역동적으로 설계된 synthetic dataset에서 강조된다.

\figref{fig:fig5} provides a qualitative comparison of the results. In the outputs from other methods, the text printed on the boxes appears with a different depth from the surface it is on, making the text distinguishable in the depth image. This occurs because optimization is performed solely on RGB rendering without considering object-level consistency, often leading to local minima in sparse-view settings. In contrast, our method benefits from instance-aware supervision, preventing 3D Gaussians from falling into such local minima and ensuring that they are properly aligned with the surfaces of the boxes. Regarding next best view selection, our approach captures not only the top surfaces and larger objects but also occluded regions and smaller objects, demonstrating a balanced exploration capability and better reconstruction.

% fig 5에서 정성적으로 결과를 비교할 수 있다. 다른 방법들에서의 결과를 보면 박스의 글씨가 그 글씨가 써져 있는 면과 depth가 달라 depth image에서도 글씨가 구분되는 것을 볼 수 있는데, 이는 optimization이 rgb rendering으로만 이뤄져서 같은 물체임을 고려하지 않아 sparse-view에서 자주 발생하는 local minima에 빠졌기 때문이다. 반면, 우리의 방법에서는 semantic supervision 덕에 3D gaussian들이 local minima에 쉽게 빠지지 않고 박스의 면에 따라 똑바르게 분포하여 이런 현상이 관찰되지 않는다. Next best view에 관해서도 우리의 방법이 물체의 윗부분이나 큰 물체 등의 exploitation에만 집중하지 않고 아랫부분이나 가려진 작은 물체 등도 잘 capture하는 것을 확인할 수 있다.

\begin{table}[t]
\centering

\caption{Ablation study for design choices.}       
\vspace{-2mm}
\label{tab:ablation}

\resizebox{\columnwidth}{!}{
\renewcommand{\arraystretch}{1.2}
{\large
\begin{tabular}{ll|cc|cc|cc}
% \hline
\toprule
 &
  \multicolumn{1}{l|}{} &
  \multicolumn{2}{c|}{\textbf{BlenderProc Dataset}} &
  \multicolumn{2}{c|}{\textbf{GraspNet Dataset}} &
  \multicolumn{2}{c}{\textbf{Captured Scenes}} \\
 &
  \multicolumn{1}{l|}{} &
  PSNR $\uparrow$ &
  D-MAE $\downarrow$ &
  PSNR $\uparrow$ &
  D-MAE $\downarrow$ &
  PSNR $\uparrow$ &
  D-MAE $\downarrow$ \\ 
  % \hline
  \midrule
\multicolumn{2}{l|}{FisherRF with mask} &
  25.34 &
  0.0695 &
  21.23 &
  0.0768 &
  15.25 &
  0.0913 \\
\multicolumn{1}{r}{} &
  + Object vector &
  25.60 &
  0.0189 &
  21.44 &
  0.0486 &
  16.05 &
  0.0583 \\
\multicolumn{1}{r}{} &
  + Confidence score &
  26.84 &
  0.0143 &
  21.39 &
  0.0488 &
  16.29 &
  0.0521 \\
\multicolumn{1}{r}{} &
  \textbf{+ IG scaling (Ours)} &
  \textbf{26.92} &
  \textbf{0.0144} &
  \textbf{21.37} &
  \textbf{0.0491} &
  \textbf{16.34} &
  \textbf{0.0507} \\ 
  % \hline
  \midrule
\multicolumn{2}{l|}{NBS (Depth) \cite{strong2024next}} &
  26.00 &
  0.0675 &
  21.21 &
  0.0765 &
  15.62 &
  0.0936 \\
\multicolumn{2}{l|}{NBS (RGB + D) \cite{strong2024next}} &
  25.83 &
  0.0675 &
  21.16 &
  0.0763 &
  15.17 &
  0.0896 \\ 
  % \hline
  \bottomrule
\end{tabular}
}
}
\vspace{-4mm}
\end{table}

\subsubsection{Ablation Study}

\tabref{tab:ablation} provides an in-depth analysis of each design choice within the proposed method. Incorporating one-hot object vector yields more accurate reconstruction without requiring additional views. Compared to FisherRF results obtained with background-removed images, the improvement is particularly pronounced in depth accuracy which decreases significantly, highlighting the necessity of semantic information for reliable robotic manipulation.

% tab 3은 propose된 pipeline에서 각 design choice들에 대하여 심층적으로 분석한 것이다. object vector의 추가는 같은 수의 view로도 더 정확한 reconstruction이 되는 것을 보여준다. background가 제거된 이미지를 사용한 fisherrf의 결과와 비교할 때 psnr 보다도 크게 줄어든 depth 정확도에서 드러나며, manipulation을 위해 semantic 정보가 필수적이라는 사실을 뒷받침한다. 

Furthermore, we observe that scaling with confidence scores derived from the one-hot object vector and opacity, combined with the information gain from training views, is also effective. While Next Best Sense \cite{strong2024next} (NBS in table) heuristically balances RGB and depth contributions by defining $IG(T,c) = IG_{rgb}(T,c) + 10 IG_{d}(T,c)$, our approach adaptively determines these weights based on the training views, leading to improved robustness.

% 또한 one-hot object vector와 opacity로부터 얻은 confidence score와 training view의 information gain을 이용한 scaling도 효과가 있음을 알 수 있다. Next best sense는 rgb와 depth rendering으로부터 얻은 Information gain을 임의로 scaling하여 $IG(T,c) = 0.1 IG_{rgb}(T,c) + IG_{d}(T,c)$ 를 사용하는데, 우리는 Training view를 이용하여 scaling을 adaptively 하기 때문에 다양한 scene에 대해 더 robust하다.

\begin{table}[ht]
\centering

\caption{Object-centric reconstruction results}       
\vspace{-2mm}
\label{tab:object_wise}

\resizebox{\columnwidth}{!}{
{\large
\renewcommand{\arraystretch}{1.2}
\begin{tabular}{c|c|cccccc}
\hline
\toprule
\multirow{3}{*}{\textbf{Scene}} &
  \multirow{3}{*}{\textbf{Target}} &
  \multicolumn{6}{c}{\textbf{Depth MAE} $\downarrow$} \\ \cline{3-8} 
 &
   &
  \multicolumn{4}{c|}{\textbf{Corner objs}} &
  \multicolumn{1}{c|}{\multirow{2}{*}{\textbf{\begin{tabular}[c]{@{}c@{}}Center\\ objs\end{tabular}}}} &
  \multirow{2}{*}{\textbf{\begin{tabular}[c]{@{}c@{}}Total\\ objs\end{tabular}}} \\ \cline{3-6}
 &
   &
  \multicolumn{1}{c|}{NW} &
  \multicolumn{1}{c|}{NE} &
  \multicolumn{1}{c|}{SW} &
  \multicolumn{1}{c|}{SE} &
  \multicolumn{1}{c|}{} &
   \\ 
   \hline
   \midrule
% \multirow{3}{*}{\begin{tabular}[c]{@{}c@{}}BlenderProc\\ Synthetic\\ Scene\end{tabular}} &
\multirow{3}{*}{\begin{tabular}[c]{@{}c@{}}Syn.\end{tabular}} &
  Whole &
  \multicolumn{1}{c|}{0.0180} &
  \multicolumn{1}{c|}{0.0365} &
  \multicolumn{1}{c|}{0.0130} &
  \multicolumn{1}{c|}{0.0215} &
  \multicolumn{1}{c|}{0.0162} &
  \textbf{0.0193} \\
 &
  Corner &
  \multicolumn{1}{c|}{\textbf{0.0132}} &
  \multicolumn{1}{c|}{\textbf{0.0124}} &
  \multicolumn{1}{c|}{\textbf{0.0114}} &
  \multicolumn{1}{c|}{\textbf{0.0166}} &
  \multicolumn{1}{c|}{0.0149} &
  0.0205 \\
 &
  Center &
  \multicolumn{1}{c|}{0.0182} &
  \multicolumn{1}{c|}{0.0503} &
  \multicolumn{1}{c|}{0.0151} &
  \multicolumn{1}{c|}{0.0286} &
  \multicolumn{1}{c|}{\textbf{0.0148}} &
  0.0225 \\
  % \hline
  \midrule
% \multirow{3}{*}{\begin{tabular}[c]{@{}c@{}}Captured\\ Real\\ Scene\end{tabular}} &
\multirow{3}{*}{\begin{tabular}[c]{@{}c@{}}Real\end{tabular}} &
  Whole &
  \multicolumn{1}{c|}{0.0621} &
  \multicolumn{1}{c|}{0.2632} &
  \multicolumn{1}{c|}{0.0695} &
  \multicolumn{1}{c|}{0.0777} &
  \multicolumn{1}{c|}{0.0334} &
  \textbf{0.0517} \\
 &
  Corner &
  \multicolumn{1}{c|}{\textbf{0.0490}} &
  \multicolumn{1}{c|}{0.1963} &
  \multicolumn{1}{c|}{\textbf{0.0624}} &
  \multicolumn{1}{c|}{\textbf{0.0619}} &
  \multicolumn{1}{c|}{0.0313} &
  0.0540 \\
 &
  Center &
  \multicolumn{1}{c|}{0.1000} &
  \multicolumn{1}{c|}{\textbf{0.1816}} &
  \multicolumn{1}{c|}{0.0683} &
  \multicolumn{1}{c|}{0.0647} &
  \multicolumn{1}{c|}{\textbf{0.0304}} &
  0.0535 \\ 
  % \hline
  \bottomrule
\end{tabular}
}
}
\vspace{-4mm}
\end{table}

\subsection{Next Best View for Object-centric Reconstruction}

% \subsubsection{Object-centric Reconstruction Evaluation}

Our method has the capability to select views that are specifically tailored to a designated target object. To clearly validate this effect, we select four corner objects in each scene that are spatially distant from one another as shown in \figref{fig:object centric recon}. This choice is intentional, as performing object-centric reconstruction on objects nearby center would likely result in overlapping visibility across most candidate views. 
%, thereby diminishing the distinctiveness and effectiveness of targeted view selection.

% 우리의 방법은 집중하고 싶은 물체가 있을 때 그 물체에 특화된 view를 select 할 수 있는 aspect를 가지고 있고, 이를 table 4에서 확인할 수 있다. 그 효과를 확실하게 확인하기 위해 각 Scene에서 서로 가장 멀리 떨어져있는 네 개의 물체를 골랐는데, 이는 서로 가까운 물체들에 대해 object-wise reconstruction을 하면 대부분의 candidate view에서 함께 보일 것이기 때문에 효과가 크지 않을 것이기 때문이다. 

For comparison, we consider three setups: (1) using the entire scene as the reconstruction target, (2) targeting non-selected center objects, and (3) individually targeting each of the four distant corner objects. This comparison highlights how our object-wise \ac{NBV} strategy adapts to different reconstruction goals and spatial distributions.

% Scene 1은 물체 간의 occlustion이 강한 좀 더 clutter된 환경에서, 2는 한 view에 scene의 일부만이 보일 수 있는 좀 더 scattered된 환경에서의 효과를 알 수 있다. 비교 대상은 reconstruction target만을 바꿔서 따로 target object를 정하지 않고 전체 scene에 대해서 했을 때, 떨어져 있는 네 물체가 아닌 나머지 물체 중 하나를 중심으로 했을 때, 그리고 네 물체 각각에 대해 따로 target으로 삼을 때를 비교했다.

% \subsubsection{Evaluation and Analysis}

As shown in \tabref{tab:object_wise} and \figref{fig:object centric recon}, when \ac{NBV} planning is targeted for a specific object rather than the entire scene, the selected viewpoints are concentrated on the corresponding object, leading to improved depth accuracy for the object.

In manipulation tasks, where the target object and its surrounding regions are substantially more critical than distant areas, this strategy not only prevents unnecessary time spent on scanning irrelevant regions but also provides a decisive advantage in both efficiency and accuracy. Consequently, it represents a highly promising direction for achieving robust and task-oriented robotic manipulation.

% \tabref{tab:object_wise}와 \figref{fig:object centric recon}에서 볼 수 있듯이 전체 씬을 목표로 NBV를 했을 때와 특정 물체를 목표로 NBV를 했을 때 그 물체에 대한 view가 집중적으로 뽑혀 depth 결과가 더 좋음을 확인할 수 있다. manipulation을 할 때 target object와 그 주변의 중요성이 멀리 떨어져 있는 부분보다 높다는 것을 생각하면, 이는 관계 없는 영역을 scan하느라 낭비되는 시간을 절약하면서 manipulation의 정확도를 높일 수 있는 아주 좋은 방법이다.

% \begin{figure}[h!]
%     \centering
%     \begin{subfigure}{1.0\columnwidth}
%         \centering
%         \includegraphics[width=\columnwidth]{figure/fig7_1_ver3.pdf}
%         \vspace{-5.5mm}
%         \caption{Convergence analysis on BlenderProc synthetic scene}
%         \label{syn}
%     \end{subfigure}
%     \begin{subfigure}{1.0\columnwidth}
%         \centering
%         % \fbox{\rule{0pt}{0.6\columnwidth} \rule{1.0\columnwidth}{0pt}}
%         \includegraphics[width=\columnwidth]{figure/fig7_2_ver3.pdf}
%         \vspace{-5.5mm}
%         \caption{Convergence analysis on captured real scene}
%         \label{real}
%     \end{subfigure}
%     \caption{PSNR and Depth MAE as the number of views increases.}
%     \label{fig:convergence}
%     \vspace{-5mm}
% \end{figure}

\begin{figure}[ht]
  \centering
  \includegraphics[width=\linewidth]{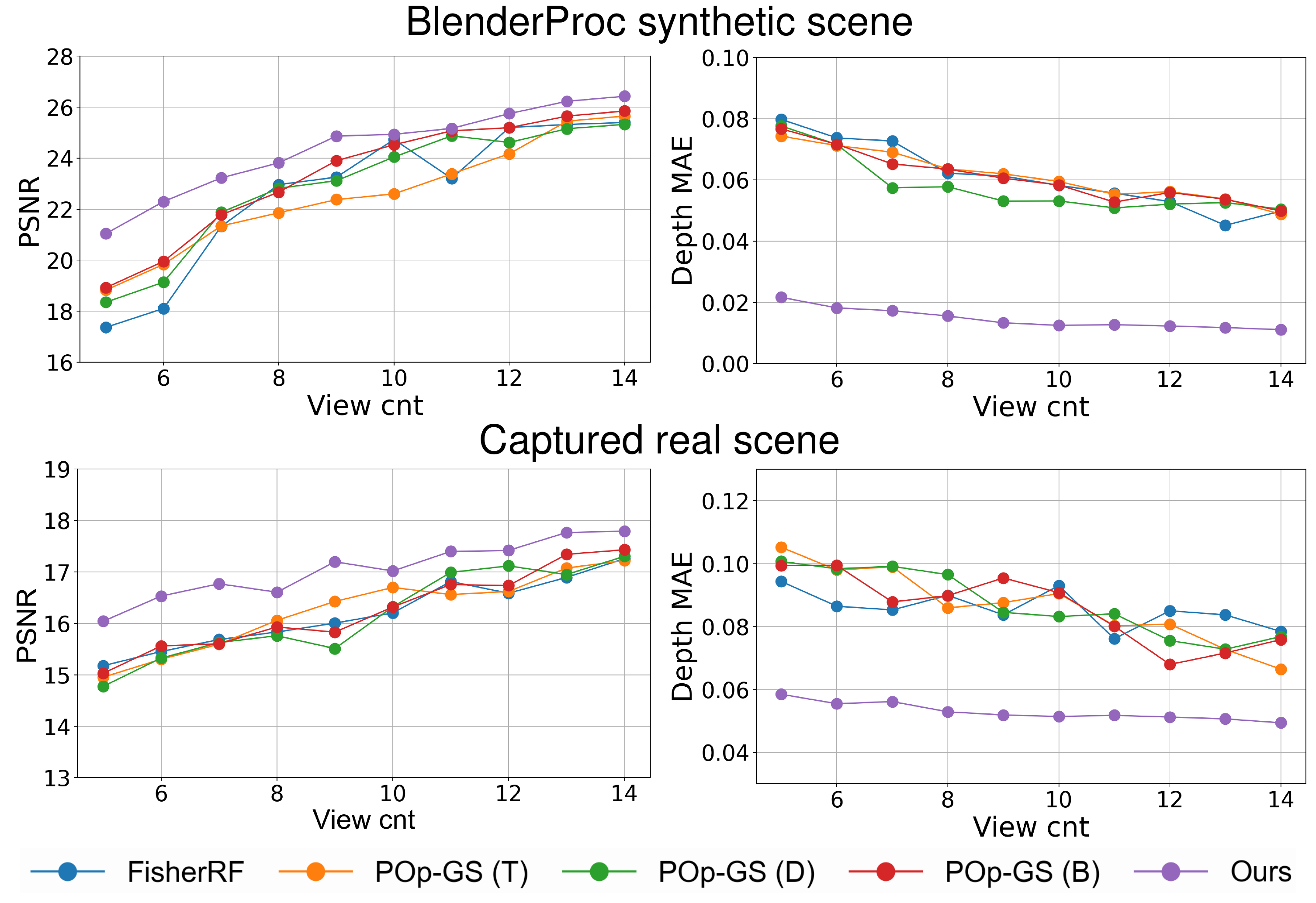}
  % \fbox{\rule{0pt}{0.3\columnwidth} \rule{0.9\columnwidth}{0pt}}
  \caption{PSNR and Depth MAE as the number of views increases from 5 to 15 in both synthetic and real scenes.}
  \label{fig:convergence}
  \vspace{-5mm}
\end{figure}

\begin{figure}[!t]
  \centering
  % \fbox{\rule{0pt}{0.4\columnwidth} \rule{0.9\columnwidth}{0pt}}
  \includegraphics[width=\columnwidth]{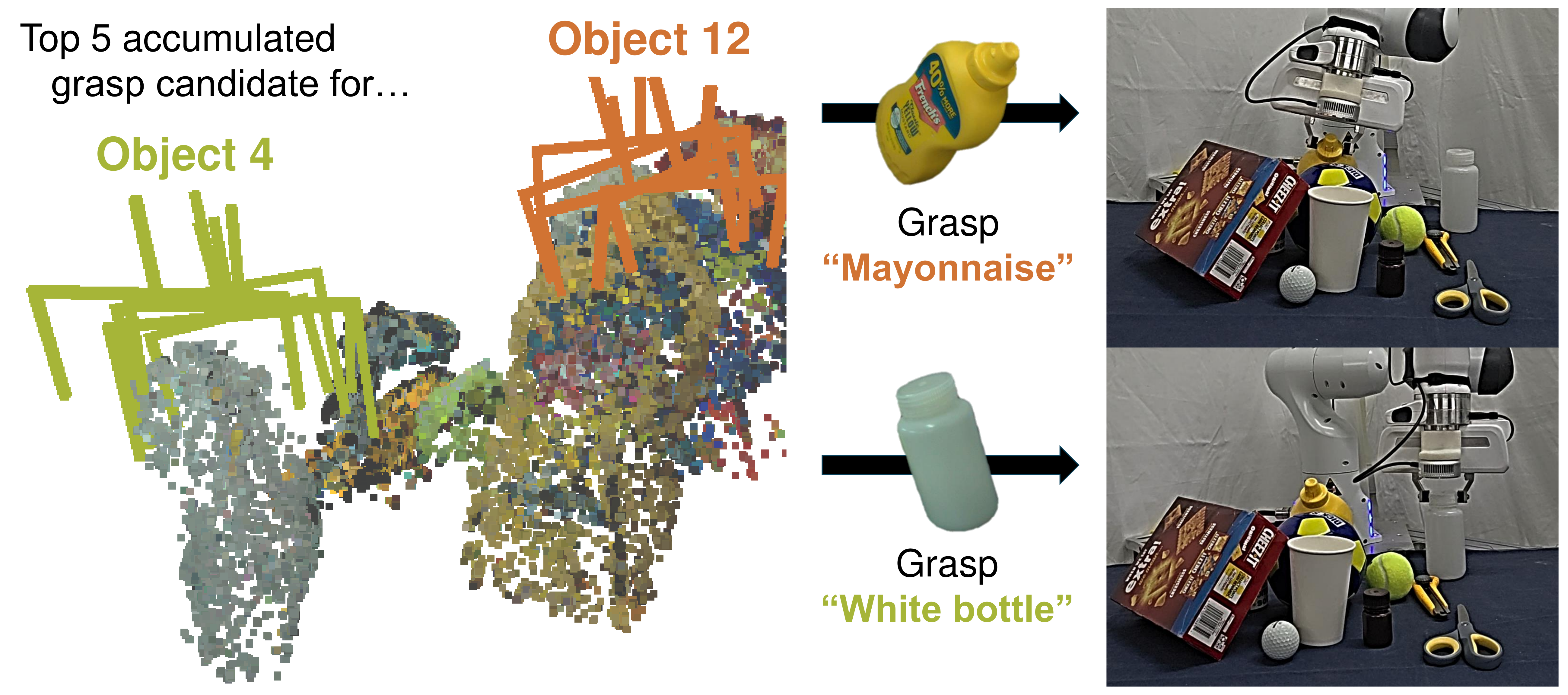}
  \caption{Left shows the top-5 aggregated grasp candidates from multi-view EconomicGrasp predictions, grouped by object. Target object can be selected using CLIP embedding similarity between a text prompt and object masks. Right shows execution snapshots of successful grasps on two target objects.}
  % 왼쪽 부분은 select된 view로부터 얻은 grasping point들을 합친 뒤 object별로 구분한 것이다. coordinate는 robot base를 의미한다. 오른쪽 부분은 그 중 두 물체를 로봇이 성공적으로 집는 것을 보여준다.
  \label{fig:grasping}
  \vspace{-5mm}
\end{figure}

\subsection{Analysis on Convergence Speed of Reconstruction}

\figref{fig:convergence} illustrates how the error with respect to the ground-truth depth decreases as the number of selected views increases, comparing our method with the baselines. 
% During the incremental acquisition of views, we perform $100v$ optimization iterations with spherical harmonics of degree 0. After the complete set of views has been obtained, the reconstruction is refined through 3000 iterations with spherical harmonics of degree 3. 
Only the final number of selected views is varied, ranging from 5 to 14, and the corresponding depth MAE is measured.

% \figref{fig:convergence}는 baseline들과 우리의 방식으로 NBV를 했을 때 GT Depth와의 오차가 view 개수에 따라 얼마나 빠르게 감소하는 지 그린 것이다. 뷰의 개수가 v일 때 정해진 수가 모이기 전에는 sh degree 0으로 100v의 iteration으로, 모인 뒤에는 sh degree 3으로 3000 iteration으로 최종 reconstruction을 하는 것을 유지하고 최종 개수만 5개부터 14개로 바꿔서 depth MAE를 확인하였다.

As shown in the graph, our method yields better initial results and ultimately achieves higher PSNR and lower depth MAE than all baselines as views increase.
% As shown in the graph, our method converges more rapidly than the baselines as the number of views increases, and it ultimately achieves higher accuracy. 
% This effect becomes even more pronounced in the case of object-centric reconstruction. 
By leveraging an appropriate termination criterion or empirical tuning, the scanning process can be terminated earlier while still maintaining reconstruction quality, thereby significantly reducing the time required for robotic operation.

% 그래프를 보면 개수가 늘어남에 따라 우리의 방법론이 baseline 보다 빠르게 수렴하고 최종적으로도 더 좋은 정확도를 가진다. 또한, object-centric reconstruction을 했을 때는 이런 현상이 더더욱 강하게 발생한다. 이는 적절한 termination criterion이나 경험적 tuning을 통해 scan 과정을 조기 종료하고 reconstructin을 진행하여 로봇이 작업하는 시간을 크게 줄일 수 있음을 시사한다.

% \textcolor{blue}{\lipsum[23]\lipsum[23]\lipsum[23]}

\subsection{Extension to Real-world Manipulation}
We apply our method to real-world robotic manipulation. To obtain grasping point, we employ EconomicGrasp \cite{wu2024economic}, and the robot platform consists of a Franka Emika Panda arm equipped with an Intel RealSense L515 camera.

% 이 섹션에서는 우리의 방법론을 실제 로봇팔 manipulation에 적용한 것이다. Grasping에는 GraspNet model을 사용하였고, view를 얻고 manipulation을 수행하는 로봇팔은 realsense L515가 mount된 Franka Emika Panda를 이용하였다. view에 대한 object mask는 mipnerf360 데이터셋에 대해서 한 것처럼 grounding sam2를 이용한 pseudo ground truth를 사용하였다.

For multi-view grasping, grasping points are first predicted by using the depth maps rendered from the Gaussian maps of each view into the EconomicGrasp model. The resulting grasp candidates are then transformed and aggregated with the camera poses obtained from the robot arm. After transforming all grasp candidates into the world coordinate frame, we filter them to retain only those with feasible positions and orientations that the robot can safely approach.

As shown in \figref{fig:grasping}, our NBV system successfully selects informative viewpoints and yields reliable grasping points from multiple views. This multi-view strategy is more robust than relying on a single predefined top-down view. In cluttered scenes, top-down views often suffer from severe occlusions, which limit the model’s ability to infer valid grasping points. In contrast, the views selected by our policy provide improved visibility of target objects, making them more suitable for accurate grasp prediction.

Furthermore, when a language prompt specifying a target object is given, we use CLIP to compute the cosine similarity between the prompt embedding and the embeddings of the object masks. The object with the highest similarity is selected, and the NBV planning is performed with a focus on that object. This demonstrates that our object-wise reconstruction framework is well suited for integration with open-vocabulary vision-language models.

% As shown in \figref{fig:grasping}, our NBV system successfully selects informative viewpoints and also returns reliable grasping points from multiple views. This approach is more robust than simply obtaining the grasping point from a top-down view. Furthermore, when a language prompt specifying a target object is given, we use CLIP to compute the cosine similarity between the prompt embedding and the embeddings of the object masks. The object with the highest similarity is selected, and the NBV planning is performed with a focus on that object. This demonstrates that our object-wise reconstruction framework is well suited for integration with open-vocabulary vision-language models.

% fig에 따르면 NBV가 적절한 view를 선택하고 그렇게 뽑은 view에서 grasping point를 얻어 가장 reliable한 graspoing point를 얻음을 알 수 있다. 이는 단순히 graspoing point를 top view에서 얻는 것보다 robust하다. 또한 물체에 대한 prompt가 주어질 때 clip을 이용하여 어떤 object mask와 가장 비슷한지를 embedding의 cosine similarity를 이용하여 찾고 그 object를 위주로 NBV를 수행하는 것을 볼 수 있다. 이는 object-wise reconstruction은 open vocabulary의 활용을 위한 vlm 모델과 결합하기도 적합함을 의미한다.

In real-world scenarios, such as bin picking or retrieving objects from the containers, the reasonable viewpoints are often physically constrained. We recommend watching the supplementary video showcasing examples where our method is applied to such restricted environments.

% 특히, real-world scenario에는 bin picking이나 상자 안에 들어있는 물체들을 보는 등 물체를 볼 수 있는 view 자체가 한정적인 경우가 많다. supplementary로 제공된 비디오에서 상자 안, 책장 안 등에 대해 NBV가 되는 것을 소개했으니 보는 것을 추천한다.

\section{Conclusion}
\label{sec:conclusion}

We presented an instance-aware \ac{3DGS}-based, confidence-aware \ac{NBV} policy that injects object masks into the \ac{3DGS} pipeline, prioritizes underexplored regions, and extends naturally to object-centric \ac{NBV}. By balancing the contributions of RGB, depth, and mask rendering, the selected views form a compact, informative set, which also yields clearer and more reliable grasping points in cluttered scenes.
% 우리는 object mask가 주어질 때 이를 3DGS framework에 넣고 못 본 region에 대한 exploration에 유리한 confidence-aware Next best view를 제안하고 이를 object-centric NBV로 까지 확장하였다. RGB, depth, mask rendering에 대한 기여도가 적절히 balancing되어 선택된 이미지들은 scene을 잘 압축하고 더 명확한 grasping point를 얻는 데에도 도움이 되었다.

A key limitation is the reliance on object masks. Although we observed good robustness with pseudo ground-truth on challenging real scenes, inaccuracies in the masks can propagate to both reconstruction and \ac{NBV} decisions. Future work will focus on online refinement of object mask via \ac{3DGS} or 3D foundation models as new views are acquired. We also plan to generate candidate viewpoints actively and to adopt principled termination criteria to enable early stopping.
% 그러나, 우리는 object mask에 크게 의존한다는 문제를 가지고 있다. 강한 clutter가 있는 real-world scene에서 pseudo-gt mask로 여전히 유효함은 확인하였으나, supervision이 정확하지 않을수록 그로 인한 역효과가 커질 것이다. 후속 연구는 3dgs나 3d foundation model 등을 통해 object mask를 robust하게 구하거나 view를 뽑을 때마다 refine하는 방법에 집중할 것이다. 또한, NBV의 candidate를 직접 생성하고 termination criterion에 따라 early-stopping이 가능한 방향으로도 확장할 계획이다.

% \newpage

% \input{ver4_finalsub/0_abstract}
% \input{ver4_finalsub/1_introduction}
% \input{ver4_finalsub/2_relatedwork}
% \input{ver4_finalsub/3_method}
% \input{ver4_finalsub/4_experiment}
% \input{ver4_finalsub/5_conclusion}

% \newpage
% \newpage

%\section*{ACKNOWLEDGMENT}
% \balance
\small
\bibliographystyle{IEEEtranN} %citeauthor
\bibliography{string-short,references}

\end{document}